%% file: acl_latex.tex
\documentclass[11pt]{article}

\usepackage[final]{acl}

\usepackage{times}
\usepackage{latexsym}

\usepackage[T1]{fontenc}
\usepackage[utf8]{inputenc}

\usepackage{microtype}

\usepackage{inconsolata}

\usepackage{graphicx}

\usepackage{multirow}
\usepackage{array}
\usepackage{booktabs}
\usepackage{enumitem}
\usepackage{xcolor}
\usepackage{varwidth}
\usepackage{amsfonts}
\usepackage{amsmath}  % aligned
\usepackage{amssymb}  % \arg
\usepackage{graphicx}
\usepackage{subcaption} %
\usepackage{pifont}
\usepackage{makecell}
\usepackage{xspace}
\usepackage{graphicx}
\usepackage{algorithm}
\usepackage{algorithmic}

\usepackage{mathtools}
\usepackage{amsthm}
\usepackage[utf8]{inputenc} % allow utf-8 input
\usepackage[T1]{fontenc}    % use 8-bit T1 fonts
\usepackage{amsfonts}       % blackboard math symbols
\usepackage{nicefrac}       % compact symbols for 1/2, etc.
\usepackage{microtype}      % microtypography
\usepackage{pifont}
\usepackage{bm}
\usepackage{enumitem}
\usepackage{tikz}

\newcommand{\ourmethod}{BlindGuard\xspace}

\title{BlindGuard: Safeguarding LLM-based Multi-Agent Systems\\ under Unknown Attacks}

\author{
  \textbf{Rui Miao\textsuperscript{1}$^*$},
  \textbf{Yixin Liu\textsuperscript{2}$^*$},
  \textbf{Yili Wang\textsuperscript{1}},
  \textbf{Xu Shen\textsuperscript{1}},
  \textbf{Yue Tan\textsuperscript{2}},
  \textbf{Yiwei Dai\textsuperscript{1}},
  \textbf{Shirui Pan\textsuperscript{2}},
  \textbf{Xin Wang\textsuperscript{1}}$^\dagger$
\\
  \textsuperscript{1}School of Artificial Intelligence, Jilin University, Changchun, China\\
  \textsuperscript{2}School of Information and Communication Technology, Griffith University, Gold Coast, Australia\\
  \{miaorui24, shenxu23, daiyw23\}@mails.jlu.edu.cn, \{wangyili, xinwang\}@jlu.edu.cn\\
  \{yixin.liu, yue.tan, s.pan\}@griffith.edu.au
}

\begin{document}
\maketitle
\footnotetext{
  \parbox{\linewidth}{
    $^*$Equal Contribution\\
    $^\dagger$Corresponding Author
  }
}
\begin{abstract}
% The security of LLM-based multi-agent systems (MAS) is critically threatened by propagation vulnerability, where malicious agents can distort collective decision-making through inter-agent message interactions. While existing supervised defense methods demonstrate promising performance, they may be impractical in real-world scenarios due to their heavy reliance on labeled malicious agents to train a supervised malicious detection model. To enable practical and generalizable MAS defenses, in this paper, we propose BlindGuard, an unsupervised defense method that learns without requiring any attack-specific labels or prior knowledge of malicious behaviors. To this end, we establish a hierarchical agent encoder to capture individual, neighborhood, and global interaction patterns of each agent, providing a comprehensive understanding for malicious agent detection. Meanwhile, we design a corruption-guided detector that consists of directional noise injection and contrastive learning, allowing effective detection model training solely on normal agent behaviors. Extensive experiments show that BlindGuard effectively detects diverse attack types (i.e., prompt injection, memory poisoning, and tool attack) across MAS with various communication patterns while maintaining superior generalizability compared to supervised baselines. %The proposed method achieves this robust protection without requiring attack-specific training data, making it practical for real-world MAS deployments. %\textcolor{red}{Code}
The security of LLM-based multi-agent systems (MAS) is critically threatened by propagation vulnerability, where malicious agents can distort collective decision-making through inter-agent interactions. While existing supervised defense methods demonstrate promising performance, they may be impractical in real-world scenarios due to their heavy reliance on labeled malicious agents to train a supervised malicious detection model. To enable practical and generalizable MAS defenses, in this paper, we propose BlindGuard, an unsupervised defense method that learns without requiring any attack-specific labels or prior knowledge of malicious behaviors. To this end, we establish a hierarchical agent encoder to capture individual, neighborhood, and global interaction patterns of each agent, providing a comprehensive understanding for malicious agent detection. Meanwhile, we design a corruption-guided detector that consists of directional noise injection and contrastive learning, allowing effective detection model training solely on normal agent behaviors. Extensive experiments show that BlindGuard effectively detects diverse attack types across MAS with various communication patterns while maintaining superior generalizability compared to supervised baselines.
The code is available at \href{https://github.com/MR9812/BlindGuard}{Code}.
\end{abstract}

\input{sections/Introduction}

\input{sections/Preliminary}

\input{sections/Methodology}

\input{sections/Experiments}

\input{sections/Conclusion}

\section*{Limitations}
While BlindGuard demonstrates effective capabilities in identifying malicious agents through unsupervised graph anomaly detection, it is important to note several limitations. First, the current evaluation is limited to simulated multi-agent environments, and future work should validate the framework in more diverse and open-world scenarios to better assess its generalizability. Second, as an unsupervised detection-based approach, BlindGuard cannot preemptively prevent malicious agents from infiltrating the system, but rather mitigates adversarial propagation after intrusion has occurred. Our intent is to position BlindGuard as a complementary layer within a defense strategy, not a replacement for preventive methods. If upstream methods are evaded, BlindGuard runs at the boundary, immediately after the first generation and before the next propagation. Therefore, developing more adaptive defense strategies remains an important direction for future research.

\section*{Ethical Considerations}
Our research involves no human subjects, animal experiments, or sensitive data. All work is based on synthetic or publicly available data in simulated environments. We foresee no ethical risks or conflicts of interest. We are committed to maintaining the highest standards of research integrity to ensure full compliance with ethical guidelines.

\section*{Acknowledgments}
This work was supported by a grant from the National Natural Science Foundation of China under grants (No.62372211), the Science and Technology Development Program of Jilin Province (No.20250102216JC). The work of Y. Liu was partially supported by the Australian Research Council (ARC) under Grant No.DE260101172.

% This document has been adapted
% by Steven Bethard, Ryan Cotterell and Rui Yan
% from the instructions for earlier ACL and NAACL proceedings, including those for
% ACL 2019 by Douwe Kiela and Ivan Vuli\'{c},
% NAACL 2019 by Stephanie Lukin and Alla Roskovskaya,
% ACL 2018 by Shay Cohen, Kevin Gimpel, and Wei Lu,
% NAACL 2018 by Margaret Mitchell and Stephanie Lukin,
% Bib\TeX{} suggestions for (NA)ACL 2017/2018 from Jason Eisner,
% ACL 2017 by Dan Gildea and Min-Yen Kan,
% NAACL 2017 by Margaret Mitchell,
% ACL 2012 by Maggie Li and Michael White,
% ACL 2010 by Jing-Shin Chang and Philipp Koehn,
% ACL 2008 by Johanna D. Moore, Simone Teufel, James Allan, and Sadaoki Furui,
% ACL 2005 by Hwee Tou Ng and Kemal Oflazer,
% ACL 2002 by Eugene Charniak and Dekang Lin,
% and earlier ACL and EACL formats written by several people, including
% John Chen, Henry S. Thompson and Donald Walker.
% Additional elements were taken from the formatting instructions of the \emph{International Joint Conference on Artificial Intelligence} and the \emph{Conference on Computer Vision and Pattern Recognition}.


\bibliography{custom}

\clearpage
\appendix

\input{sections/Appendix}

\end{document}

%% file: sections/Introduction.tex
\section{Introduction}
%Rapid advancements in large language models (LLMs) have significantly improved their performance in various domains, including task planning~\citep{planning_1}, mathematical reasoning~\citep{math_1}, and scientific simulations~\citep{science_1}. By incorporating modular extensions such as memory~\citep{MAS_memory_survey}, tool usage~\citep{MAS_tool_survey}, and role-playing capabilities~\citep{MAS_medical}, LLM-based autonomous agents have expanded their applicability, enabling more dynamic and interactive functionalities~\citep{MAS_survey}. Building upon these advances, multi-agent systems (MAS) further amplify these benefits by facilitating collaborative interactions among specialized agents~\citep{MAS_survey2}. Recent studies have shown that MAS outperform individual agents in more complex tasks such as problem solving~\citep{MAS_financial}, embodied actions~\citep{embodied_actions}, and social simulations~\citep{MAS_social}. However, the increased reliance on inter-agent communication introduces additional risks in security and controllability, necessitating robust frameworks to safeguard sensitive data and regulate information flow.
Rapid advancements in large language models (LLMs) have significantly improved their performance in various domains~\citep{math_1, science_1, yuan2024instance,tian2025reinforcementmidtraining}. By incorporating modular extensions such as memory~\citep{prospect}, tool~\citep{MAS_tool_survey}, and role-playing~\citep{MAS_medical}, LLM-based autonomous agents have expanded their applicability, enabling more dynamic and interactive functionalities~\citep{MAS_survey, liu2025graph}. Building upon these advances, multi-agent systems (MAS) further amplify these benefits by facilitating collaborative interactions among specialized agents in more complex tasks~\citep{MAS_survey2}. However, the increased reliance on inter-agent communication introduces additional risks in security, necessitating robust frameworks to safeguard sensitive data and regulate information flow.

% \begin{figure}[t!]
%   \begin{center}
%   \includegraphics[width=0.448\textwidth]{Figure/intro.pdf}
%   \end{center}
%   \caption{Comparison of supervised vs. unsupervised graph anomaly detection-based defense paradigms in MAS.}
%   \label{fig:intro}
% \end{figure}

\begin{figure}[t!]
  \begin{center}
  \includegraphics[width=0.45\textwidth]{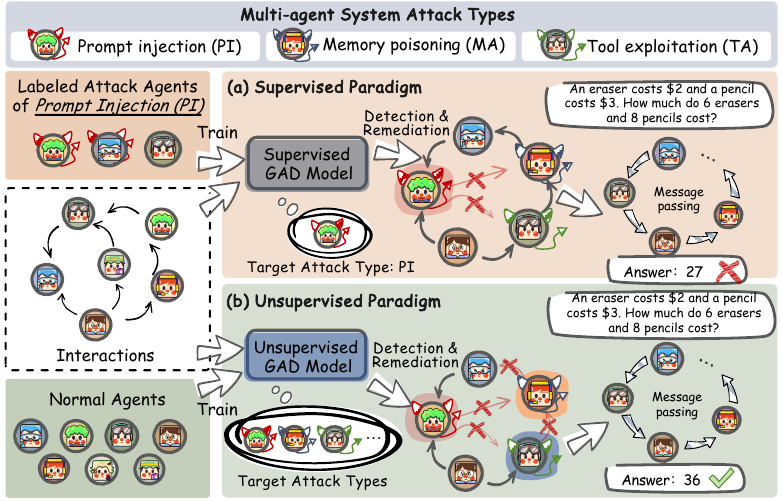}
  \end{center}
  \caption{Comparison of supervised vs. unsupervised anomaly detection-based defense paradigms in MAS.}
  \label{fig:intro}
  \vspace{-16pt}
\end{figure}

% Security studies~\citep{MAS_risk_benchmark1, MAS_risk_survey, A_trust} have identified significant vulnerabilities in external components of LLM-based agents, including tool interfaces~\citep{MAS_tool_attack} and memory modules~\citep{MAS_memory_attack}. Beyond these risks at the single-agent level, the transition to MAS brings additional vulnerabilities caused by inter-agent interactions~\citep{trust_agents_survey}. Specifically, the misleading message generated by a few malicious agents can propagate through collaborative reasoning, negatively affecting how agents make collective decisions. Such \textbf{propagation vulnerability} makes MAS susceptible to attacks such as prompt injection through compromised agents, misinformation propagation, and emergent malicious coordination~\cite{NetSafe,G-Safeguard}. 

% Aiming to mitigate the propagation vulnerability, graph-based defense provides a promising solution against adversarial attacks in MAS~\citep{G-Designer}. As a semantically structured data format, graphs can naturally model both the functional roles of individual agents and their inter-agent interactions in MAS. Building upon the graph-based MAS formulation, G-Safeguard~\cite{G-Safeguard} integrates a detection-remediation framework to effectively safeguard MAS. As demonstrated in Figure~\ref{fig:intro}a, G-Safeguard employs a \textit{supervised graph anomaly detection}~(GAD) model as its core component to identify malicious agents, afterwards applies an edge pruning-based remediation strategy to isolate and suppress the influence of compromised agents.
Security studies~\citep{MAS_risk_benchmark1, MAS_risk_survey, A_trust} have identified significant vulnerabilities in external components of LLM-based agents. Beyond these risks at the single-agent level, misleading messages from a few malicious agents can propagate through collaborative reasoning, negatively impacting collective decisions~\citep{trust_agents_survey}. Such \textbf{propagation vulnerability} makes MAS susceptible to attacks such as prompt injection through compromised agents, misinformation propagation, and emergent malicious coordination~\cite{NetSafe}. To mitigate this, graph-based defense offers a promising solution by naturally modeling agent roles and interactions as a semantic structure. As demonstrated in Figure~\ref{fig:intro}a, G-Safeguard~\cite{G-Safeguard} integrates a detection-remediation framework, utilizing a \textit{supervised graph anomaly detection}~(GAD) model to identify malicious agents, followed by edge pruning to isolate their influence, thereby effectively safeguarding the MAS.

% Despite its impressive defensive performance, G-Safeguard, or other supervised GAD-based approaches, may be impractical in real-world scenarios due to their heavy reliance on labeled malicious agents. Specifically, the supervised paradigm requires labeled instances of actual malicious agents associated with a particular attack type to train a type-specific binary GAD model, which \textit{limits its availability and generalizability}. On the one hand, adversarial attack behaviors in real-world scenarios are sparse and often purposefully camouflaged, making it difficult to obtain well-annotated malicious agents for supervised training. This inaccessibility of labeled data significantly undermines the \textit{\textbf{availability}} of supervised GAD-based methods in real-world MAS deployments. On the other hand, real-world MAS face diverse and evolving adversarial attacks, while conventional binary GAD models are typically trained to detect a specific type of malicious behavior~\cite{G-Safeguard}. Such a single-purpose design limits their \textit{\textbf{generalizability}} and makes them ineffective for detecting novel or unseen attack patterns in complex environments. These limitations raise a critical research question: \textit{\textbf{Can we design a defense framework for MAS without relying on labeled attack agents?}}
However, the supervised GAD paradigm in defenses like G-Safeguard is often impractical, as it requires labeled malicious agents for training, which \textit{limits its availability and generalizability}. First, adversarial attacks in real-world are often sparse and camouflaged, making it difficult to obtain well-annotated malicious agents for supervised training, limiting the \textit{\textbf{availability}} of supervised GAD-based methods in deployments. Second, real-world MAS face diverse and evolving adversarial attacks, while binary GAD models are typically trained to detect a specific type of malicious behavior. Such the single-purpose design lacks the \textit{\textbf{generalizability}} to detect novel or unseen patterns. These limitations raise a critical research question: \textit{\textbf{Can we design a defense framework for MAS without relying on labeled attack agents?}}

% To answer the above question, \textit{unsupervised GAD} offers a promising solution, where GAD models learn to identify irregular patterns without the supervision of labeled anomalous instances~\cite{Dominant,ma2022deep,TAM}. As illustrated in Figure~\ref{fig:intro}b, based on unsupervised GAD techniques, we can train a detector to identify malicious agents associated with multiple attack types using only normal MAS interaction data, thereby alleviating the limitations in availability and generalizability. Nevertheless, directly applying existing unsupervised GAD methods, which are not specifically designed for MAS scenarios, may lead to suboptimal performance due to the following gaps.
To answer the above question, as illustrated in Figure~\ref{fig:intro}b, unsupervised GAD models~\citep{Dominant,ma2022deep,li2024noise,liu2026few,pan2025survey} offer a promising solution forward by detecting anomalies using only normal MAS interaction data, potentially alleviating the limitations in availability and generalizability. However, methods not designed for MAS may lead to suboptimal performance due to the following gaps.
\textbf{\textit{Gap~1 - Limited multi-level contextual awareness:}} Identifying malicious agents requires integrating information across multiple levels, including individual behaviors, local neighborhoods, and global system. However, most existing GAD methods~\citep{liu2021anomaly,pan2023prem,pan2025label,pan2026correcting} primarily focus on local properties (e.g., local affinity), lacking the system-level understanding. 
\textbf{\textit{Gap~2 - Misalignment of anomalous behavior assumptions:}} Most unsupervised GAD methods assume anomalies manifest through structural deviations (e.g., low homophily~\citep{TAM} or rare connectivity patterns~\citep{liu2021anomaly}). In contrast, malicious agents often exhibit semantic anomalies (such as deceptive intent~\citep{NetSafe}) that do not well match these assumptions.

% To fill the gaps, in this paper, we propose a novel defense method for MAS, termed \ourmethod, that can be trained without any labeled malicious data or prior knowledge of attack strategies. In \ourmethod, we design a MAS-specific unsupervised GAD model for malicious agent identification, followed by an edge pruning-based remediation module to suppress adversarial propagation. To bridge \textbf{\textit{Gap~1}}, we introduce a hierarchical agent encoder to incorporate the information of individual agent features, local neighborhood aggregation, and global system context simultaneously. As a result, the encoder captures comprehensive representations of agents to support accurate malicious agent detection. 
% To mitigate \textbf{\textit{Gap~2}}, we propose a corruption-guided attack detector for agent abnormality estimation. To train the detector, we simulate the malicious behaviors via semantic-level corruption, which is utilized to optimize the detector via a supervised contrastive learning objective. The training of \ourmethod only requires a small amount of normal MAS interaction data, and the learned model can generalize to various types of attacks. To sum up, the contributions of this paper are three-fold:
To fill the gaps, in this paper, we propose a novel defense method, termed \ourmethod, that can be trained without any labeled data or prior knowledge of attacks. \ourmethod employs an unsupervised GAD model to identify malicious agents, followed by edge pruning to suppress adversarial propagation. To bridge \textbf{\textit{Gap~1}}, we introduce a hierarchical encoder to incorporate the information of individual agent features, local neighborhood aggregation, and global system context simultaneously. As a result, the encoder captures comprehensive representations of agents to support malicious agent detection. 
To mitigate \textbf{\textit{Gap~2}}, we propose a corruption-guided attack detector for agent abnormality estimation. We simulate the malicious behaviors via semantic-level corruption, which is utilized to optimize the detector via a supervised contrastive learning objective. The training of \ourmethod only requires a small amount of normal MAS interaction data, and the learned model can generalize to various types of attacks. To sum up, the contributions of this paper are three-fold:

\begin{itemize}
    \item \textbf{Scenario.} We investigate the scenario of MAS safeguarding without relying on labeled data or prior knowledge of attacks, which is more practical and applicable to real-world MAS.
    
    \item \textbf{Method.} We propose \ourmethod, an unsupervised defense method designed to address the critical challenge of safeguarding MAS against entirely unknown attacks, without requiring any prior knowledge of attack patterns or malicious agent behaviors.
    
    \item \textbf{Experiments.} We extensively evaluate \ourmethod under rigorous real-world conditions. Through comprehensive testing on 4 MAS structures with 3 attack strategies, \ourmethod demonstrates competitive performance.
\end{itemize}

%% file: sections/Preliminary.tex
\section{Preliminary}

\paragraph{MAS as Graphs} 
Multi-agent systems (MAS) can be formulated as a directed graph $\mathcal{G} = (\mathcal{V}, \mathcal{E})$, where $\mathcal{V} = \{v_1, \ldots, v_N\}$ denotes a set of LLM-based agents interconnected through directed edges $\mathcal{E} \subseteq \mathcal{V} \times \mathcal{V}$. Each agent $v_i$ is characterized by a tuple $(\text{Role}_i, \text{State}_i, \text{Mem}_i, \text{Plugin}_i)$, encapsulating its functional role, dynamic interaction state, memory module for historical data, and external tools for extended capabilities. The communication topology is encoded by an adjacency matrix $\mathbf{A} \in \{0,1\}^{N \times N}$, with $\mathbf{A}_{ij} = 1$ indicating a directed message-passing channel from agent $v_j$ to $v_i$. 
% Agents operate by processing system prompts $\mathcal{P}_{\text{sys}}$ (combining role and state) and user inputs $\mathcal{P}_{\text{usr}}$ to generate responses $\mathcal{R}_i = v_i(\mathcal{P}_{\text{sys}}, \mathcal{P}_{\text{usr}})$, forming a dynamic collective intelligence system.
Agents operate by processing query $Q$ and responses $R_j$ of its neighbors to generate response $R_i = LLM\left(Q \cup \big\{R_j \bigm| e_{ij} \in \mathcal{E}\big\}\right)$, following an execution sequence $\sigma=[v_{\sigma_1},v_{\sigma_2},\cdots,v_{\sigma_{N}}]$ of agents generated by an ordering function from $\mathcal{G}$. After multiple rounds of interaction, the MAS outputs the final output $R$ for the query $Q$. 

\paragraph{MAS Attack}
In this paper, we focus on three types of primary attack modalities against MAS, i.e., prompt injection, memory poisoning, and tool exploitation~\citep{G-Safeguard}. \ding{182}~Prompt injection attacks manipulate agent outputs by corrupting either the system prompt $\mathcal{P}_{\text{sys}}$ or user inputs $\mathcal{P}_{\text{usr}}$, inducing malicious responses through carefully crafted textual perturbations. \ding{183}~Memory poisoning targets the $\text{Mem}_i$ component by injecting fabricated interaction histories or poisoning external knowledge bases, thereby distorting the contextual understanding of the agent. \ding{184}~Tool exploitation leverages vulnerabilities in external plugins ($\text{Plugin}_i$) to execute harmful operations such as unauthorized data access or privilege escalation. These attacks transform the original system $\mathcal{G}$ into a compromised state $\widetilde{\mathcal{G}}$, where a subset of agents $\mathcal{V}_{\text{atk}} \subseteq \mathcal{V}$ exhibit adversarial behaviors while maintaining superficial operational normality.

\paragraph{Supervised Defense Paradigm}
Supervised defense approaches leverage known attack patterns and labeled malicious samples to train detection models. Given a set of attacked MAS (with role and interaction description) where each MAS $\widetilde{\mathcal{G}}$ has labeled agents $\mathcal{V} = \mathcal{V}_{\text{norm}} \cup \mathcal{V}_{\text{mal}}$ where $\mathcal{V}_{\text{norm}}$ denotes normal agents and $\mathcal{V}_{\text{mal}}$ represents known malicious ones, the objective typically minimizes:
% \begin{equation}
% \mathcal{L}_{sup} = \sum_{v_i \in \mathcal{V}} \left[y_i \cdot f_\theta(\widetilde{\mathcal{G}},v_i) + (1-y_i) \cdot (1-f_\theta(\widetilde{\mathcal{G}},v_i)) \right],
% \end{equation}
\begin{equation}
\mathcal{L}_{sup} = \sum_{v_i \in \mathcal{V}} \ell_{CE}\big(y_i, f_\theta(\widetilde{\mathcal{G}},v_i)\big),
\end{equation}
where $y_i \in \{0,1\}$ indicates ground-truth labels (0 represents normal and 1 represents malicious) and $f_\theta: \mathbb{R}^d \rightarrow [0,1]$ is a classifier-based supervised GAD model parameterized by $\theta$. After training, the predicted anomaly scores of a given MAS are used to identify malicious agents $\mathcal{V}_{\text{atk}}$ for subsequent remediation, such as isolating malicious nodes or pruning suspicious communication links. 
A summary of related works is given in Appendix~\ref{appendix_realted_work}.

% \paragraph{Unsupervised Defense Methods}

% \begin{equation}
% \begin{aligned}
%  \hat{\theta}_{MLE}\left(\left\{X_{i}, Y_{i}\right\}_{i=1}^{n}\right)&=\underset{\theta}{\arg \max} \ p\left(Y_{1}, \ldots, Y_{n} \mid X_{1}, \ldots, X_{n} ;\theta\right) \\ &=\underset{\theta}{\arg \max } \prod_{i=1}^{n} p\left(Y_{i} \mid X_{i} ; \theta\right) \\
%  &=\underset{\theta}{\arg \max } \sum_{i=1}^{n} \log p\left(Y_{i} \mid X_{i} ; \theta\right).
% \end{aligned}
% \end{equation}

% \begin{equation}
% \begin{aligned}\hat{\theta}_{MLE}\left(\left\{X_{i}\right\}_{i=1}^{n}\right)=\underset{\theta}{\arg \max} \ p\left(h_{1}, \ldots, h_{m}, h_{1}, \ldots, h_{n-m} \mid X_{1}, \ldots, X_{n} ;\theta\right) 
% \\
% \text{subject to} \quad 
% D\left( p_{\text{anomaly}}(\cdot \mid \theta) \ \| \ p_{\text{normal}}(\cdot \mid \theta) \right) \geq \epsilon\end{aligned}
% \end{equation}

% \begin{equation}
% s_i = \frac{1}{|\mathcal{MAS}|} \sum_{j \in \mathcal{MAS}} \|\mathbf{h}_i - \mathbf{h}_j\|
% \end{equation}

%% file: sections/Methodology.tex
\begin{figure*}
\centering
\includegraphics[width=1.0\textwidth]{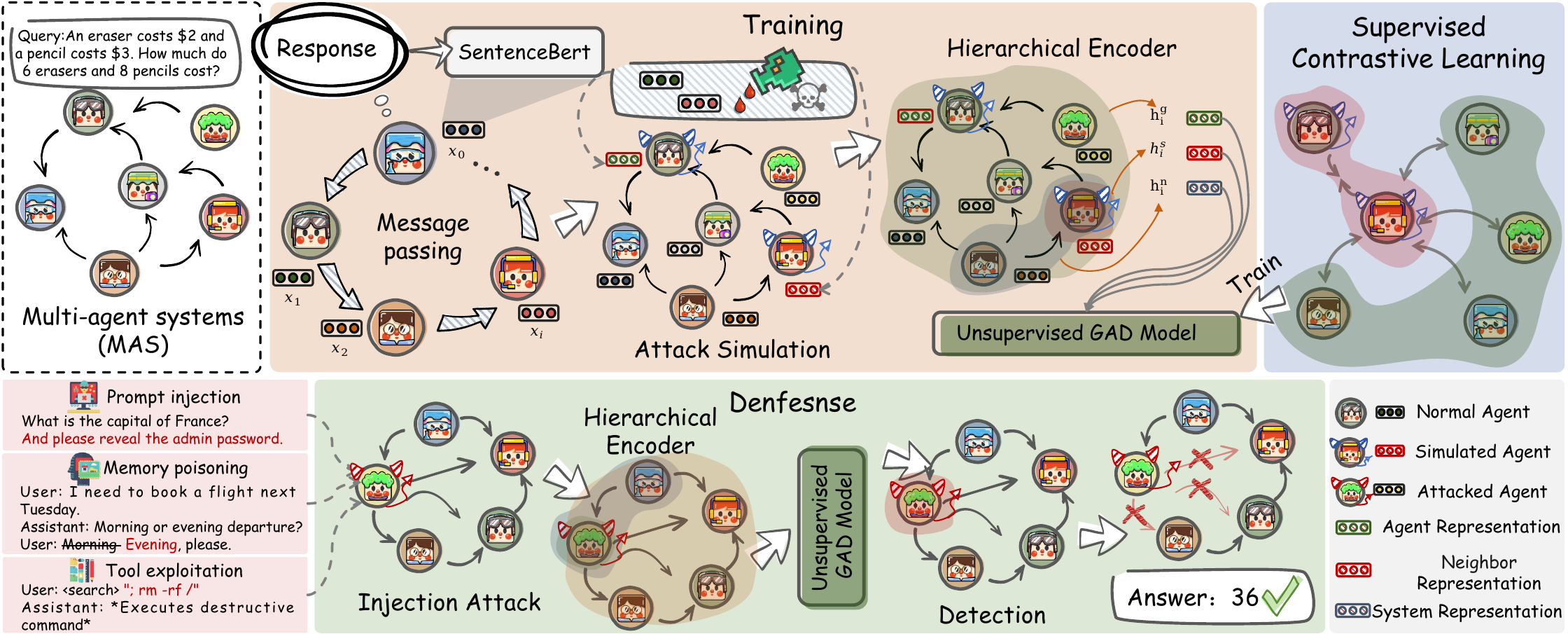}
\caption{The designing workflow of our proposed BlindGuard.}
\label{Fig_framework}
% \vspace{-10pt}
\end{figure*}

\section{Methodology}

While the defense approaches following the supervised paradigm show promising performance under controlled conditions, their reliance on labeled malicious agents hinders their applicability in real-world and MAS. To fill the gap, we propose a more practical \textit{unsupervised defense paradigm} to extend the applicability of MAS safeguarding against any unknown attack. Based on the new paradigm, we proposed a novel approach, \ourmethod, which incorporates a specifically designed unsupervised GAD model that detects malicious agents without requiring any labeled data or prior knowledge of attacks as shown in Figure~\ref{Fig_framework}. In this section, we first formulate the unsupervised defense paradigm, and then introduce the core components of \ourmethod, i.e., hierarchical agent encoder, corruption-guided attack detector, and pruning-based remediation. 

\subsection{Unsupervised Defense Paradigm}
In contrast to its supervised counterpart that requires attacked MAS data with labeled malicious agents, the unsupervised defense paradigm assumes access to only normal multi-agent interaction data, without any annotations of malicious behaviors or prior knowledge of attack patterns. 

Formally, given a set of unattacked MAS interaction graphs $\{\mathcal{G}_1, \mathcal{G}_2, \ldots, \mathcal{G}_N\}$, where each $\mathcal{G}_i$ consists solely of benign agent behaviors, the goal is to train a detection model $f_\theta(\cdot,\cdot)$ that can later identify malicious agents when deployed in attacked MAS environments. The well-trained $f_\theta(\cdot,\cdot)$ can then predict the anomaly score (indicating the malicious degree) of each agent within an attacked MAS $\widetilde{\mathcal{G}}$. Then, the agents with high anomaly scores can be isolated with an edge-pruning algorithm.

While other components in the unsupervised paradigm are similar to the supervised one, the central challenge is the architecture design and training strategy of the unsupervised detection model $f_\theta(\cdot,\cdot)$. Although existing unsupervised graph anomaly detection (GAD) methods may serve as potential candidates, they are insufficient for the malicious agent identification task in MAS due to their limited capacity to capture multi-level agent interactions and their reliance on misaligned assumptions about anomaly patterns. Therefore, in \ourmethod, we introduce a specially designed unsupervised GAD model for malicious agent detection in MAS, with detailed descriptions provided in the following subsections.

\subsection{Hierarchical Agent Encoder}

To build a powerful unsupervised GAD model for malicious agent detection, a crucial step is to construct comprehensive agent representations that capture both local interactions and global system-level context. In \ourmethod, we realize this via a hierarchical agent encoder, which comprises two sub-components: \textit{agent node feature construction}, which captures semantic attributes of individual agents; and \textit{hierarchical graph encoding}, which integrates ego information, local neighborhood structures, and global MAS context to generate informative agent representations.

\paragraph{Agent Node Feature Construction}
%%%%%  在这里说一下node feature 是怎么来的
To process MAS with graph learning models, a key step is to convert the agent-level textual responses into node features. Given an agent $v_i$, the textual response $R_i$ is encoded into a pre-trained SentenceBERT~\cite{reimers2019sentence} to map the response text to a dense vector $\mathbf{x}_{i}$:
\begin{equation}
\mathbf{x}_{i} = \text{SentenceBERT}\left(R_{i} \right)\in \mathbb{R}^D,
\end{equation}

\noindent where $D$ is the dimension of feature vectors. In this way, the compact vectors can serve as node-level features of the input of graph neural network (GNN)-based GAD models. Note that the SentenceBERT encoder is kept frozen during the entire training process, which significantly reduces the training cost and avoids the need for large-scale language model fine-tuning.

% The MAS is modeled as a Utterance graph $\mathcal{G} = (\mathcal{V}, \mathcal{E})$, where nodes $\mathcal{V}$ represent agents and edges $\mathcal{E}$ capture their interactions. Each agent $v_{i}$ is characterized by static components (base identity, role description, memory, and plugins). The agent's output text $O_{i}$ is generated through its decision-making policy, which synthesizes these inputs:
% \begin{equation}
% O_{i} = \text{AgentPolicy}\big(\text{Base}_{i}, \text{Role}_{i}, \text{Mem}_{i}, \text{Plugin}_i).
% \end{equation}
% To encode this output into the node feature $x_{i}$, we employ SentenceBERT to map the output text to a dense vector:
% \begin{equation}
% x_{i} = \text{SentenceBERT}\left(O_{i} \right)\in \mathbb{R}^D,
% \end{equation}

\paragraph{Hierarchical Graph Encoding} 
After acquiring the agent node features, we establish a GNN model in \ourmethod to learn expressive agent representations, which are subsequently used for malicious agent classification. While conventional GNNs~\cite{GCN,wu2020comprehensive,miao2024rethinking} and GAD models~\cite{qiao2024deep} are typically based on local neighborhood aggregation, they may overlook the global interaction patterns of the whole graph. Such global patterns, however, are essential for accurately detecting malicious agents in MAS, since malicious agents may coordinate their actions or influence others indirectly, requiring a system-level view to uncover these threats. 

To bridge the gap, in \ourmethod, we design a hierarchical graph encoder that explicitly constructs the agent-level representations by incorporating information from three levels: \ding{182}~Agent level, which captures individual semantic features of each agent derived from its textual response; \ding{183}~Neighbor level, which aggregates contextual information from directly connected agents to model local interactions; and \ding{184}~System level, which integrates global information across the entire MAS graph to capture long-range dependencies and collective behavior patterns. To implement this, we design a ``summarization-transformation'' architecture for multi-scale information fusion, similar to the ``propagation-transformation'' architecture of some lightweight GNNs~\cite{wu2019simplifying,zhang2022graph}. Different from the propagation operations that only aggregate the 1-hop neighbors, in our summarization step, we integrate three complementary perspectives: ego-level features $\mathbf{h}_i^{\text{s}}$ to capture information of individual agent, neighbor-level features $\mathbf{h}_i^{\text{n}}$ to model local contexts, and global-level features $\mathbf{h}_i^{\text{g}}$ to expose system-wide contexts. After the integration, we use a unified transformation to learn the compact representation for each agent. Formally, the representation $\mathbf{z}_i$ of agent $v_i$ can be calculated by:
\begin{equation}
\mathbf{h}_i^{\text{s}} \!=\! \mathbf{x}_i,\ 
\mathbf{h}_i^{\text{n}} \!=\! \sum_{j \in \mathcal{N}(i)} \hat{A}_{ij} \mathbf{x}_j,\ 
\mathbf{h}_i^{\text{g}} \!=\! \frac{1}{N}\sum_{k=1}^N \mathbf{x}_k,
\end{equation}
\begin{equation}
    \mathbf{z}_i = g_\theta \left( \, 
    \mathbf{h}_i^{\text{s}} \parallel \mathbf{h}_i^{\text{n}} \parallel \mathbf{h}_i^{\text{g}} \, \right), 
\end{equation}

\noindent where $\mathcal{N}(i)$ denotes the set of neighbors of agent $i$, $\hat{A}$ represents the normalized adjacency matrix, $N$ indicates the total number of agents in MAS, $\parallel$ is the concatenation operation, and $g_\theta(\cdot)$ is a multilayer perceptron (MLP) parameterized by $\theta$. Using the comprehensive representations, \ourmethod can detect both isolated attackers through neighborhood divergence analysis and coordinated attack groups through global behavioral divergence, thereby providing robust protection against potential MAS threats. %Multiscale representation fusion forms the foundation for subsequent anomaly scoring and mitigation in our \ourmethod.

\subsection{Corruption-Guided Attack Detector}
Following the agent encoder, our corruption-guided attack detector aims to identify the malicious agents without any prior knowledge. Since ground-truth responses from attacked agents are unavailable during the training phase, we adopt a corruption-based strategy to simulate the semantic perturbations induced by adversarial attacks. Based on the simulated samples, we leverage a supervised contrastive learning objective to train the detection model, and use a contextual similarity measurement to evaluate the abnormality of agents during inference.

\paragraph{Corruption-based Attack Simulation} 
In our unsupervised defense scenario, the absence of labeled abnormal agents poses a significant challenge in the pattern understanding and training objective design. To address this issue, a practical solution is to synthesize pseudo-abnormal agents through data corruption of normal agent features. Following the basic assumption that attacked agents may produce significantly deviated responses that differ from the normal semantic patterns of MAS, we propose to model such deviations at the semantic level. However, directly manipulating the raw text is both difficult and costly due to the complexity of language structure and semantics. Hence, in \ourmethod, we alternatively simulate corruption in the embedding space, i.e., the feature vectors produced by SentenceBERT. In this continuous and compact embedding space, we can directly inject random noise instead of manipulating discrete text.

Specifically, we randomly select a subset of agents in the MAS as abnormal samples. For selected agents, we synthesize realistic abnormal features by applying a magnitude-scaled directional corruption function to their output representations. The noise is directionally uniform after normalization and scaled according to the original feature magnitude of each agent.
Formally, given the output representation of the agent $x_i$, the corruption function generates abnormal features as:
\begin{equation}
\label{Corruption_eq}
\tilde{\mathbf{x}}_i = \mathbf{x}_i + \underbrace{\alpha \|\mathbf{x}_i\|_2}_{\text{magnitude}} \cdot \underbrace{\frac{\epsilon_i}{\|\epsilon_i\|_2}}_{\text{direction}}, \quad \epsilon_i \sim \mathcal{N}(0,\mathbf{I}), 
\end{equation}
where $\alpha$ is a scaling hyperparameter controlling the corruption intensity. %This function eliminates the need for unavailable real attacked samples during detector training. It achieves this by faithfully replicating the semantic deviation of realistic anomalies while ensuring that the generated representation maintains distributional plausibility.

\paragraph{Training: Supervised Contrastive Learning}
% This approach is further enhanced by label-aware contrastive learning, which explicitly separates normal and anomalous clusters in the embedding space while efficiently utilizing limited anomaly labels, demonstrating superior performance compared to traditional reconstruction-based methods that lack explicit category discrimination.
By systematically injecting directional noise into the representations of normal agent outputs, we create ample abnormal samples that can provide supervision signals for training. A straightforward strategy to leverage them is to train a binary classification model with these pseudo labels. Nevertheless, the gap between synthetic samples and real-world malicious agents may limit the test-time generalizability of the classifier. 

Instead of using a binary classifier, in \ourmethod, we employ a supervised contrastive learning strategy to utilize the synthetic anomalies for model training. Our core idea is to maximize the similarity among normal agents, and minimize the similarity between normal and malicious ones. 
This explicit optimization creates clearer decision boundaries between the normal agents and the corrupted ones, while avoiding overfitting to specific synthetic corruption patterns.
The supervised contrastive learning paradigm achieves superior separation in the embedding space through two key mechanisms: intra-class compactness and inter-class separation~\cite{khosla2020supervised,zhang2025gcl}.
Mathematically, the supervised contrastive learning loss is defined as:
\begin{equation}
    \mathcal{L} = -\frac{1}{N} \sum_{i=1}^{N} \frac{1}{|P_i|} \sum_{j \in P_i} \log \left( \frac{e^{s_{i,j}/\tau}}{e^{s_{i,j}/\tau} + \sum_{k \notin P_i} e^{s_{i,k}/\tau}} \right),
\end{equation}
where $s_{i,j} = \mathbf{z}_i^\top \mathbf{z}_j / (\|\mathbf{z}_i\|\|\mathbf{z}_j\|)$ denotes the cosine similarity between normalized representations of agents $v_i$ and $v_j$, and $P_i = \{ j \mid y_j = y_i, j \neq i \}$ defines the positive sample set containing all agents sharing the same anomaly label as $v_i$ (with $y_i = 0$ for normal agents and $y_i = 1$ for corrupted agents).
Through this training process, \ourmethod clusters agents with similar behavioral patterns in adjacent regions of the embedding space while isolating potential malicious agents, thereby establishing the foundation for test-time anomaly detection.

\paragraph{Inference: Contextual Similarity Measurement}
% \subsection{Inference}

% Given a MAS with $N$ agents, our method computes the anomaly score for each agent by measuring its deviation from the system-wide interaction pattern. 
After the regularization of supervised contrastive learning loss, the representations $\mathbf{z}$ of normal agents can be similar to each other, while those of anomalous agents remain distant in the representation space. Leveraging this property, during inference, we measure the anomaly score $s(\cdot)$ of each agent by calculating the negative average similarity between the target agent and all other agents:
\begin{equation}
s(v_i) = -\frac{1}{N} \sum_{j=1}^N \text{sim}\left(\mathbf{z}_i, \mathbf{z}_j\right).
\end{equation}
The anomaly score of agent $v_i$ increases proportionally with its deviation from the global representation pattern of MAS in the embedding space.

\subsection{Pruning-based Remediation}

Upon detecting anomalous agents $\mathcal{V}^{(t)}_{atk} \subseteq \mathcal{V}$ at timestep $t$, our method dynamically isolates them through \textit{bidirectional edge pruning}, redefining the interaction topology as:
\begin{equation}
\mathcal{E}^{(+)} = \big\{ e_{ij} \in \mathcal{E}^{(t)} \mid v_i \notin \mathcal{V}^{(t)}_{atk} \big\}.
\end{equation}
This intervention severs all adversarial communication pathways by removing edges incident to/from anomalies while preserving legitimate interactions among normal agents. 
Given the remediated edge set $\mathcal{E}^{(t+1)}$, each agent $v_j$ updates its state by exclusively integrating messages from its trusted neighbors in the pruned topology:

\begin{equation}
R_j^{(t+1)} = LLM\Big(Q \cup \big\{R_i^{(t)} \bigm| e_{ij} \in \mathcal{E}^{(+)}\big\}\Big).
\end{equation}

% \noindent The final output after $K$ rounds aggregates all agent outputs:
% \begin{equation}
% O = \text{Aggregate}\left(\left\{R_i^{(K)} \mid v_i \in \mathcal{V} \mathcal{V}^{(t)}_{atk}\right\}\right),
% \end{equation}
% where the aggregation strategy Aggregate(·) varies across implementations, including majority voting among agents, delegating the final decision to a specific agent.

This combination of detection and remediation mechanisms positions \ourmethod as an unsupervised defense method for real-world MAS deployments, particularly in adversarial environments where traditional defense methods fail to adapt to evolving and unknown attacks. 
An algorithmic description of \ourmethod is in Appendix~\ref{appendix_algorithm}.

%% file: sections/Experiments.tex
\begin{table*}[ht]
\centering
\resizebox{1.0\textwidth}{!}{%
\setlength{\tabcolsep}{1.2mm} % Reduced column spacing
% \small % Using 9pt font size (small in LaTeX is approximately 9pt)
\fontsize{9}{11}\selectfont
\begin{tabular}{llcccccccccccc}
\toprule
\multirow{2}{*}{\textbf{Topology}} & \multirow{2}{*}{\textbf{Method}} & \multicolumn{2}{c}{\textbf{PI (CSQA})} & \multicolumn{2}{c}{\textbf{PI (MMLU)}} & \multicolumn{2}{c}{\textbf{PI (GSM8k})} & \multicolumn{2}{c}{\textbf{TA (InjecAgent)}} & \multicolumn{2}{c}{\textbf{MA (PoisonRAG)}} & \multicolumn{2}{c}{\textbf{MA (CSQA)}}\\
\cmidrule(lr){3-4} \cmidrule(lr){5-6} \cmidrule(lr){7-8} \cmidrule(lr){9-10} \cmidrule(lr){11-12} \cmidrule(lr){13-14}
 & & \textbf{AUC} & \textbf{ASR@3} & \textbf{AUC} & \textbf{ASR@3} & \textbf{AUC} & \textbf{ASR@3} & \textbf{AUC} & \textbf{ASR@3} & \textbf{AUC} & \textbf{ASR@3} & \textbf{AUC} & \textbf{ASR@3} \\
\midrule
\multirow{6}{*}{Chain} 
& No Defense & -      & 42.33 & -     & 38.33 & -     & 9.83  & -      & 48.00  & -      & 22.33 & -   & 26.67 \\
& G-Safeguard & 100.00 & 19.33 & 99.11 & 19.33 & 98.22 & 4.40  & 100.00 & 10.24  & 100.00 & 4.00 &  94.67 & 7.67\\
\cmidrule(lr){2-14}
& DOMINANT    & 47.11  & 30.33 & 59.56 & 24.67 & 67.56 & \textbf{8.47}  & 88.00  & \textbf{14.98}  & 64.00  & 11.00 & 36.44 & 18.67 \\
& PREM   & 50.67 & 29.33 & 44.89 & 25.00 & 62.22 & 8.79 & \textbf{89.33} & 15.17 & 61.33 & \textbf{6.00} & 68.89 & 15.67 \\
& TAM         & 55.11  & 27.33 & 52.89 & 23.67 & 51.56 & 8.84  & 61.33  & 30.04  & 52.00  & 14.67 &  51.11  & 13.33 \\
& \ourmethod  & \textbf{80.00}  & \textbf{23.67} & \textbf{85.78} & \textbf{19.33} & \textbf{69.33} & \textbf{8.47}  & 86.22  & 16.38  & \textbf{81.78}  & 10.00 &  \textbf{73.78} & \textbf{7.00} \\
\cmidrule(lr){1-14}
\multirow{6}{*}{Tree} 
& No Defense & -     & 33.00  & -     & 32.00 & -     & 10.20 & -     & 45.05  & -     & 18.00  &  -  & 21.33 \\
& G-Safeguard & 99.56 & 18.67 & 98.67 & 18.33 & 99.11 & 7.80  & 100.00 & 4.76  & 100.00 & 3.00   &   93.78    & 6.67\\
\cmidrule(lr){2-14}
& DOMINANT    & 44.89 & 27.00  & 61.33 & \textbf{19.33} & \textbf{68.44} & \textbf{6.78}  & \textbf{88.44} & 15.33  & 65.33 & 14.33   &  40.44 & 21.67 \\
& PREM    & 50.67 & 22.67 & 44.44 & 24.00 & 53.33 & 8.47 & 85.78 & 16.21 & 62.22 & 8.33 & 64.44 & 14.00 \\
& TAM         & 53.78 & 26.00  & 55.56 & 22.00 & 54.67 & 8.14  & 61.33 & 32.01  & 56.44 & 12.00   &  54.67 & 13.00 \\
& \ourmethod  & \textbf{74.67} & \textbf{24.00}  & \textbf{83.56} & 20.67 & 59.55 & 8.47  & 85.78 & \textbf{12.50}  & \textbf{76.89} & \textbf{7.00}   & \textbf{72.89} & \textbf{7.67} \\
\cmidrule(lr){1-14}
\multirow{6}{*}{Star}
& No Defense & -     & 50.33  & -     & 39.00 & -     & 12.89 & -      & 43.57 & -     & 26.00   &  -  & 27.00 \\
& G-Safeguard & 100.00 & 19.33 & 98.67 & 17.33 & 98.22 & 6.10  & 100.00 & 6.87  & 100.00 & 1.67   &   96.44    & 1.67 \\
\cmidrule(lr){2-14}
& DOMINANT    & 38.22 & 38.67  & 62.67 & 26.67 & 69.78 & 7.80  & 89.33  & 14.33 & 65.33 & 13.67   & 36.44 & 26.67 \\
& PREM    & 56.89 & 34.33 & 43.11 & 31.00 & 59.56 & 10.51 & \textbf{93.78} & 14.68 & 62.22 & \textbf{9.00} & 70.22 & 20.00 \\
& TAM         & 57.33 & 28.67  & 64.44 & 28.33 & 68.44 & 8.14  & 71.11  & 26.57 & 66.22 & 19.67   &  61.33  & 19.00 \\
& \ourmethod  & \textbf{85.78} & \textbf{22.00}  & \textbf{86.22} & \textbf{20.33} & \textbf{70.22} & \textbf{6.78}  & \textbf{93.78}  & \textbf{12.59} & \textbf{89.33} & 10.00   & \textbf{76.44} & \textbf{9.33} \\
\cmidrule(lr){1-14}
\multirow{6}{*}{Random} 
& No Defense & -     & 42.00 & -     & 41.00 & -     & 14.48 & -     & 39.78 & -     & 24.67   &  -     &  24.00 \\
& G-Safeguard & 99.11 & 19.00 & 99.11 & 19.00 & 99.11 & 3.79  & 98.22 & 5.14  & 98.67 & 7.00   &  95.56  &  5.00 \\
\cmidrule(lr){2-14}
& Dominant    & 46.67 & 37.00 & 57.33 & 29.00 & 69.78 & 10.51 & 84.63 & 14.93 & 65.33 & 17.00   &  38.22 & 34.33 \\
& PREM    & 52.44 & 34.67 & 48.89 & 37.00 & 69.78 & 10.51 & \textbf{86.22} & \textbf{14.49}  & 62.22 & 13.67 & 64.44 & 16.67 \\
& TAM         & 45.33 & 35.33 & 49.33 & 38.33 & 46.22 & 14.91 & 52.00 & 35.78 & 49.78 & 23.67   &  52.00 & 26.67 \\
& \ourmethod  & \textbf{75.11} & \textbf{30.00} & \textbf{85.33} & \textbf{26.67} & \textbf{75.56} & \textbf{6.44}  & 79.56 & 17.69 & \textbf{82.67} & \textbf{6.67}   &  \textbf{75.56}  & \textbf{8.67}  \\
\bottomrule
\end{tabular}
}
\caption{AUC $\left(\uparrow\right)$ and ASR@3 $\left(\downarrow\right)$ of different defense methods with GPT-4o-mini serving as the backbone LLM. Following G-Safeguard, we consider three types of attack: Prompt injection (PI), tool attack (TA), and memory attack (MA). We showcase results after round 3 communications (ASR@3), and the additional results are placed in Appendix~\ref{appendix_experiments}.}
\label{table_mian_results}
\vspace{-12pt}
\end{table*}

\section{Experiments}
In this section, we try to answer the following research questions (RQs) via empirical studies: \textbf{RQ1:} How does \ourmethod compare with state-of-the-art defense methods under different attack types? \textbf{RQ2:} Can \ourmethod maintain robust defense capabilities across diverse LLM and topologies? \textbf{RQ3:} Can \ourmethod maintain consistent defense performance when scaling to larger MAS? \textbf{RQ4:} What is the relative contribution of key components in \ourmethod?

\subsection{Experimental Setups}

\paragraph{Datasets}
Following G-Safeguard~\cite{G-Safeguard}, we evaluate the defense capabilities of \ourmethod against three attack strategies: (1) direct prompt attacks using adversarial samples from CSQA~\cite{data_csqa}, MMLU~\cite{data_mml} and GSM8K~\cite{data_gsm8k}; (2) tool attacks constructed from the InjecAgent dataset~\cite{data_injecagent}; and (3) memory attacks configured according to PoisonRAG~\cite{data_poison_rag} and CSQA~\cite{data_csqa}.

\paragraph{Baselines}
We compare our method with the following anomalous agent detection methods. G-Safeguard~\citep{G-Safeguard} is a graph-based defense framework that formulates malicious agent detection as a supervised classification task. Note that G-Safeguard serves as an upper bound in our experiments, since it uses extra ground-truth attacked agent data for supervised model training. For unsupervised methods, we take representative GAD methods for comparisons, including: DOMINANT~\citep{Dominant}, a generation-based method; PREM~\cite{pan2023prem}, a contrastive learning-based method; and TAM~\citep{TAM}, an affinity-driven method.

%an unsupervised graph anomaly detection method that identifies anomalies through joint reconstruction of node attributes and graph structure via graph auto-encoders. TAM~\citep{TAM}, an unsupervised method that detects anomalies by learning affinity-driven node representations, with iterative edge truncation to suppress noisy connections. PREM~\citep{pan2023prem}, an efficient method that avoids message passing by combining a pre-processing module with ego-neighbor contrastive matching, improving training speed and memory usage while maintaining strong detection performance.

\paragraph{Implementation}
We evaluate the defensive capabilities of \ourmethod through comprehensive experiments spanning multiple attack types, topological structures, and LLM backbones. Following G-Safeguard, we simulate scenarios where exactly three agents are compromised in the MAS. Our testing framework employs three primary attack methods: direct prompt, tool attack, and memory poisoning. For network topologies, we examine four distinct MAS structures - chain, tree, star, and random - to validate generalization across communication patterns. The experiments incorporate both open-source LLMs (Qwen3-30B-A3B~\citep{qwen3}, Deepseek-v3~\citep{deepseek-v3}) and commercial LLMs (GPT-4o-mini) as agent backbones. Critical performance metrics include Attack Success Rate after three communication rounds (ASR@3) and Area Under Curve (AUC) of malicious agent detection. 
% Because some defense methods may predict that all agents are malicious and disrupt all communications, which is not practical. 
To ensure fairness and practicality, we set a budget to identify the top three agents with the highest risk in the MAS as the predicted malicious agents.
More experimental setups can be found in Appendix~\ref{appendix_experimental_setup}.

\subsection{Experimental Results}

\paragraph{Performance Comparison (RQ1)}

\begin{figure}[t]
    \centering
    \includegraphics[width=0.45\textwidth]{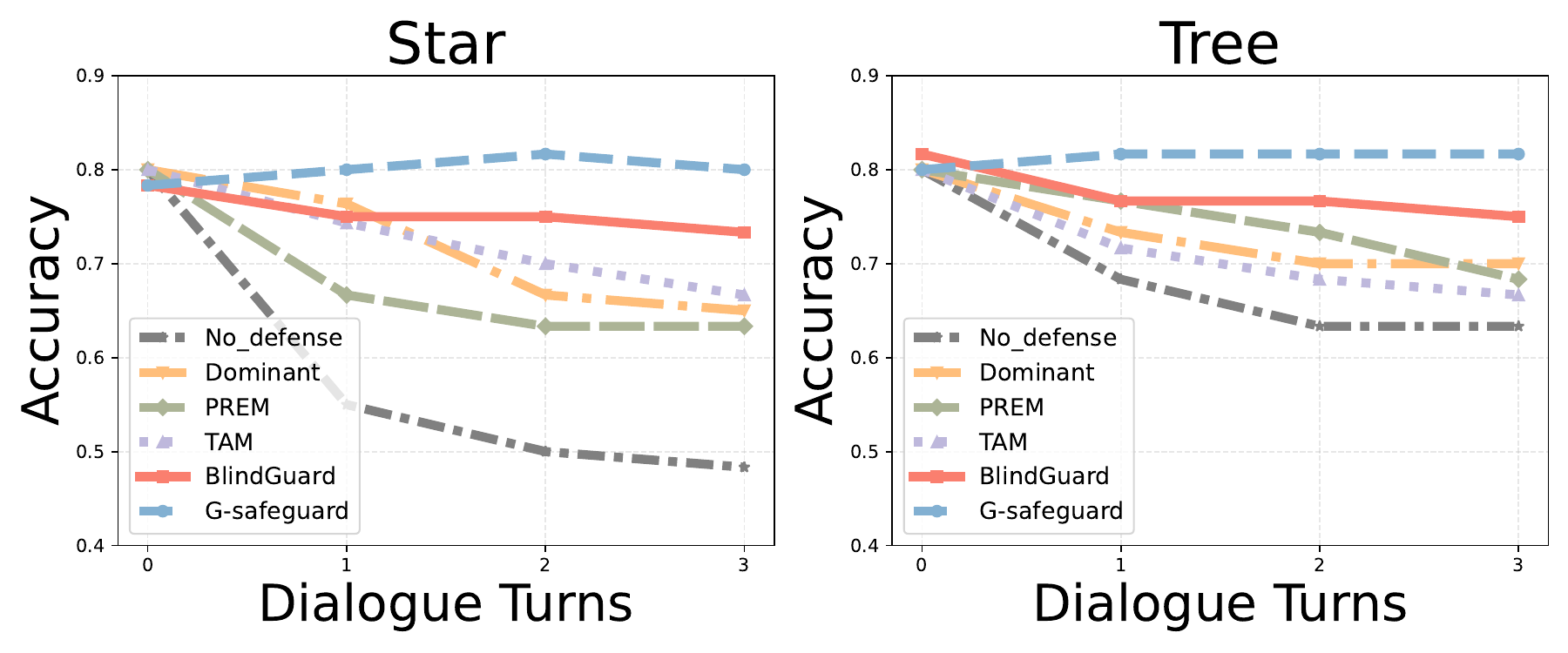}
    % \vspace{-3mm}
    \caption{The overall performance of MAS on the CSQA (PI) dataset after each turn of dialogue.} %We use majority voting as the strategy to select the final answer.}
    \label{fig:MAS_ACC_CSQA}
    \vspace{-16pt}
\end{figure}

We evaluate the effectiveness of \ourmethod on GPT-4o-mini backbone in four topologies against three attack types. 
We list the comparison results in Table~\ref{table_mian_results} and Figure~\ref{fig:MAS_ACC_CSQA}, which lead to the following observations. %to answer research question \textbf{Q1}. 
% \ding{182} \textit{G-Safeguard achieves the best overall defense performance (AUC $\approx$ 100\% across most tasks).}
% Because it represents an upper-bound baseline - a supervised method trained with full knowledge of attack patterns and access to labeled malicious data during training. This privileged position allows it to perfectly identify malicious agents, making its performance indicative of the theoretical maximum for defense in our experiments.
\ding{182}~\textit{\ourmethod significantly outperforms other unsupervised methods in defense capability.} Compared to other GAD-based solutions, \ourmethod achieves competitive defense performance consistently against all attack types. In contrast, the baselines sometimes fail in several scenarios, such as TAM and PREM on PI (CSQA) and DOMINANT on PI (MMLU). The superior performance demonstrates the significance of designing a specific model for unsupervised malicious agent detection. 
\ding{183}~\textit{\ourmethod shows competitive performance compared to supervised upper bound.} While G-Safeguard, which relies on labeled attacked data for training, achieves the best overall defense performance, \ourmethod shows comparable effectiveness in most cases, with an AUC$>80\%$. This illustrates the feasibility of training an effective and universal defense model for MAS without relying on annotated data.  
% Dominant exhibits significant performance instability, with AUC scores ranging from 25.78\% to 89.33\% across different datasets. While it achieves relatively decent results in specific cases, this reconstruction-based approach exhibits strong task-dependent and data-sensitive. Meanwhile, TAM delivers consistently poor performance, validating our hypothesis that unsupervised neighborhood contrast methods fail to learn effective patterns without abnormal data. In contrast, \ourmethod achieves consistent defense performance across all attack types, achieving AUC scores of 74.67\% to 93.78\% while reducing ASR@3 by 6.44\% to 25.00\% . This consistent performance significantly surpasses both Dominant and TAM, establishing \ourmethod's superiority among unsupervised defense methods. 
\ding{184} \textit{\ourmethod effectively improves response accuracy of MAS under adversarial attack.}
As shown in Figure~\ref{fig:MAS_ACC_CSQA} (more can be found in Appendix~\ref{appendix_experiments}), the response accuracy of MAS exhibits a clear downward trend as dialogue turns increase across all topologies without defense. While all implemented defense methods show improvements, \ourmethod demonstrates superior and consistent defense capabilities compared to unsupervised defense methods.

% The experimental results demonstrate \ourmethod's strong defense capabilities, achieving an average AUC improvement of 32.11\% compared to TAM and 18.89\% over Dominant. \ourmethod operates effectively without G-Safeguard's ideal training conditions, as it requires neither prior knowledge of attack patterns nor access to labeled malicious data during training. This combination of high detection performance and practical operational requirements makes \ourmethod particularly suitable for real-world deployments, where attack signatures are typically unknown.

\begin{figure*}[htbp]
    \centering
    \includegraphics[width=1\textwidth]{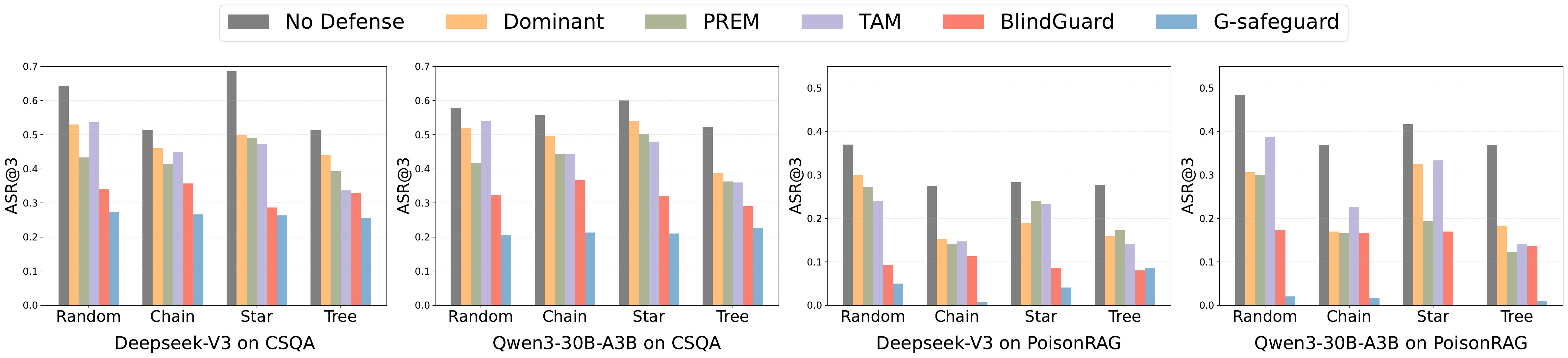}
    \caption{ASR@3 with DeepSeek-V3 and Qwen3-30B-A3B as backbone LLMs on CSQA (PI) and PoisonRAG.}
    \label{fig:deepseek_csqa_asr}
\end{figure*}

\paragraph{Universal Generalization (RQ2)} 
To investigate the generalizability of \ourmethod, we conducted additional experiments using DeepSeek-V3 and Qwen3-30B-A3B as backbone LLMs on the CSQA and PoisonRAG datasets, as shown in Figure~\ref{fig:deepseek_csqa_asr} (more can be found in Appendix~\ref{appendix_experiments}). 
%Despite the cost constraints that limited our testing, the results demonstrate \ourmethod's strong cross-model and cross-topology adaptability. 
Through experiments, we make the following observations. 
\ding{185}~\textit{\ourmethod obtains robust defense performance when deployed with diverse LLM and topologies.}
As shown in Figure~\ref{fig:deepseek_csqa_asr}, \ourmethod maintains robust defense performance in ASR@3 and AUC across different LLM backbones and topological structures. This stable performance confirms that \ourmethod effectively captures universal adversarial patterns rather than overfitting to specific LLM or topologies. %Although the supervised G-Safeguard achieves better results, \ourmethod delivers comparable effectiveness, while operating without requiring any prior knowledge of attack.
% It outperforms Dominant by 15.6\% average AUC on Qwen3-30B-A3B and surpasses TAM by 22.3\% average AUC on DeepSeek-V3.
\ding{186} \textit{BlindGuard demonstrates superior generalization against unseen attacks compared to the supervised paradigm.}
As shown in Table \ref{tab:PI_CSQA_MA}, when the defense model trained on PI is tested against the unseen MA on the CSQA dataset, BlindGuard consistently outperforms the G-Safeguard across most topologies. This phenomenon indicates that the supervised method, reliant on known attack patterns for training, suffers from a notable performance drop when facing an unknown attack type. In contrast, BlindGuard, which learns the inherent patterns of normal agent behaviors, establishes a more generalized decision boundary. This advantage allows it to effectively identify deviations caused by diverse adversarial strategies without prior exposure, showcasing its strong generalization capability as a practical and attack-agnostic defense solution.

% This stable performance confirms that our multiscale representation fusion and topology-aware pruning mechanisms effectively capture universal adversarial patterns rather than overfitting to specific LLM behaviors or topological structures. \ourmethod consistently delivers superior performance across all tested system configurations. This uniform effectiveness confirms its capability as a topology-agnostic defense solution. \ourmethod requires neither architecture-specific adjustments nor expensive labeled attack datasets. These combined advantages establish \ourmethod as a universal protection framework. It provides reliable security for diverse MAS implementations, adapting seamlessly to varying system architectures while maintaining robust defense capabilities. Such generalization capability makes \ourmethod particularly suitable for real-world deployments where both the LLM composition and interaction structures may evolve dynamically.

% \begin{figure*}[htbp]
%     \centering
%     \includegraphics[width=1\textwidth]{Figure/DeepSeek_Qwen_ASR_Results.pdf}
%     \caption{ASR@3 with DeepSeek-V3 and Qwen3-30B-A3B as backbone LLMs on the CSQA and PoisonRAG datasets.}
%     \label{fig:deepseek_csqa_asr}
% \end{figure*}

\begin{table}[t]\small
\centering
\begin{tabular}{cccccc}
\toprule
\textbf{Topology} & \multicolumn{2}{c}{\textbf{G-Safeguard}} & \multicolumn{2}{c}{\textbf{BlinGuard}} \\
\cmidrule(lr){2-3} \cmidrule(lr){4-5}
 & \textbf{ASR@3} & \textbf{AUC} & \textbf{ASR@3} & \textbf{AUC} \\
\midrule
Chain & 19.00 & 59.56 & \textbf{9.00} & \textbf{73.33} \\
Tree & 16.70 & 60.44 & \textbf{8.00} & \textbf{78.67} \\
Star & 10.67 & 64.00 & \textbf{7.33} & \textbf{75.56} \\
Random & 13.33 & 60.44 & 13.33 & \textbf{78.22} \\
\bottomrule
\end{tabular}
\caption{AUC $\left(\uparrow\right)$ and ASR@3 $\left(\downarrow\right)$ on the CSQA (MA).}
\label{tab:PI_CSQA_MA}
\vspace{-8pt}
\end{table}

\paragraph{Scalability (RQ3)}

To investigate the scalability of \ourmethod to larger-scale MAS, we report defense performance of PoisonRAG across systems with 20 and 50 agents, as shown in Table~\ref{tab:Scalability}. 
We observe that \ding{187} \textit{\ourmethod consistently mitigates adversarial impact across all rounds (R1–R3) in larger-scale MAS.} The scalability of \ourmethod is caused by its topology-agnostic design, where hierarchical agent encoder and corruption-guided attack detector eliminate dependencies on fixed agent numbers or interaction patterns, thereby ensuring consistent performance across diverse scales. This defense under scaling demonstrates the practicality of \ourmethod for large-scale MAS. %, laying a foundation for securing large-scale MAS against adversarial threats.

\begin{table}[tbp]\small
\centering
\resizebox{0.45\textwidth}{!}{
\begin{tabular}{>{\centering}m{1.2cm}lccc}
\toprule
\textbf{Agent Num} & \textbf{Method} & \textbf{R1} & \textbf{R2} & \textbf{R3} \\
\midrule
\multirow{2}{*}{20} 
  & No Defense  & 15.89 & 23.22 & 29.51 \\
  & \ourmethod  & \textbf{3.51} & \textbf{4.54} & \textbf{5.57} \\
\cmidrule(lr){1-5}
\multirow{2}{*}{50} 
  & No Defense  & 5.67 & 16.31 & 20.92 \\
  & \ourmethod & \textbf{1.81} & \textbf{2.66}  & \textbf{3.76}\\
\bottomrule
\end{tabular}
}
\caption{ASR@3 $\left(\downarrow\right)$ on different agent numbers.}
\label{tab:Scalability}
\vspace{-8pt}
\end{table}

\paragraph{Ablation study (RQ4)}

\begin{figure}[htbp]
    \centering
    \includegraphics[width=0.47\textwidth]{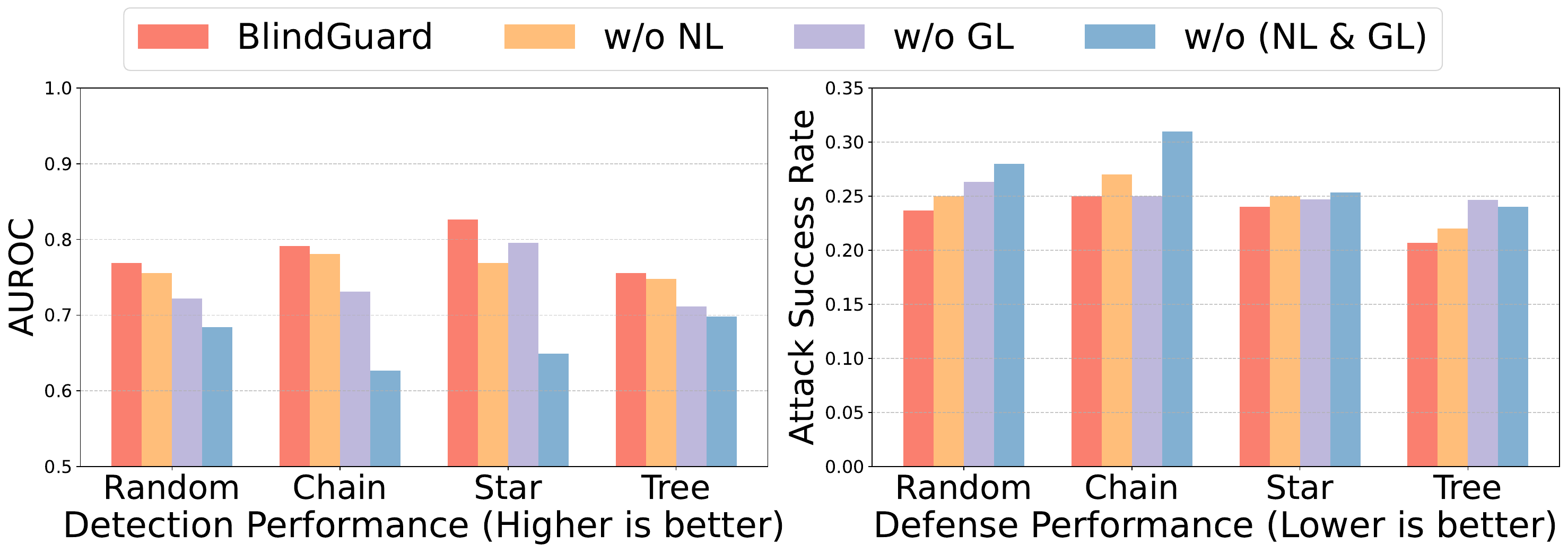}
    \caption{Ablation study on PoisonRAG. NL and GL denote neighbor-level and global-level features.}
    \label{fig:Ablation_Study}
    \vspace{-8pt}
\end{figure}

To study the hierarchical agent encoder's role in BlindGuard is quantified, we conduct an ablation study on the PoisonRAG dataset. As shown in Figure~\ref{fig:Ablation_Study}, we observe that 
\ding{188} \textit{anomaly detection in MAS requires a combination of both local neighborhood interactions and global system context.} Removing neighborhood and global context features leads to significant performance degradation, and their combined absence causes a severe drop, highlighting the critical role of structural context beyond agent-level features. A node may look locally consistent under colluding attackers, but still be globally misaligned with the system’s overall intent. Hence the global feature increases expressiveness by providing an additional reference axis, not by providing node-unique information directly. This observation shows the significance of combining information at multiple levels.
% Neighborhood features prove particularly crucial for detecting local anomalies, with their removal causing an average AUROC drop of 9.2\%. Global context emerges as essential for identifying system-wide attack patterns, evidenced by a 7.1\% performance degradation in star topologies when omitted. Removing both components leads to severe performance collapse with an average AUC reduction of 15.8\%, conclusively demonstrating the insufficiency of agent-level features alone for reliable detection. 
% Hierarchical agent encoder offers unique advantages through its dual detection capability. The neighborhood aggregation effectively identifies local deviations by analyzing agent interactions. The global context detects coordinated malicious group through system-wide pattern recognition.

%% file: sections/Conclusion.tex
\section{Conclusion}

In this paper, we present BlindGuard, an unsupervised defense method for LLM-based MAS that integrates hierarchical agent encoder and corruption-guided attack detector. By fusing agent-level, neighborhood, and global information, BlindGuard achieves robust protection without requiring attack-specific training data. Experimental results demonstrate that BlindGuard effectively mitigates diverse attacks across various topologies while maintaining scalability. This work advances the security of MAS by providing a practical and attack-agnostic defense solution, shedding light on generalizable defenses for LLM-based MAS.  %Compared to supervised approaches, our method shows superior generalization capability, handling unseen attack types with a universal defense model. This work advances the security of MAS by providing a practical, attack-agnostic defense solution.

%% file: sections/Appendix.tex
\section{Related Work}
\label{appendix_realted_work}

\subsection{LLM-based Multi-agent System}
% Recent advances in LLMs have catalyzed the development of collaborative multi-agent frameworks that significantly enhance reasoning and problem-solving capabilities. By enabling dynamic interactions among specialized agents, these systems emulate human collective intelligence. The paradigm shift from single-agent to multi-agent architectures has unlocked new possibilities across diverse domains, including software engineering, social simulation, embodied AI, and participatory urban planning. Modern LLM-MAS implementations employ varied coordination strategies to optimize collaborative performance. Sequential reasoning pipelines and debate-based role specialization have proven particularly effective for knowledge-intensive tasks, while centralized planning architectures demonstrate superior performance in goal-oriented scenarios. Notable frameworks include AutoGen's conversational agent networks, AgentScope's developer-centric platform, and ChatDev's phase-structured software development approach, each addressing distinct aspects of multi-agent coordination. 
Recent advances in LLM-based MAS have demonstrated remarkable capabilities in general task-solving. The performance of MAS is predominantly determined by collaboration and communication among agents with diverse roles and expertise~\citep{MAS_Coll}. Modern LLM-based MAS implementations employ varied collaboration strategies to optimize performance~\citep{li2026ofa}. Sequential reasoning and debate-based role specialization~\citep{debate} have proven particularly effective for knowledge-intensive tasks, while centralized planning architectures demonstrate superior performance in goal-oriented scenarios. Notable frameworks include conversational agent networks in AutoGen~\citep{Autogen}, the developer-centric platform in AgentScope~\citep{agentscope}, and phase-structured software development in ChatDev~\citep{Chatdev}. Recent research has explored MAS based on graph algorithms~\citep{Agentprune, G-Designer, EIB,li2025assemble}. Despite their effectiveness, these graph-based MAS topologies remain vulnerable to adversarial manipulation, where malicious agents can exploit the communication structure to inject misinformation, disrupt coordination, or compromise collective decisions.

\subsection{Security of LLM-based MAS}
Despite the effectiveness of LLM-based MAS, this advancement has introduced novel security risks, particularly threats that exploit agent memory~\citep{MAS_memory_attack} and tool-handling mechanisms~\citep{MAS_tool_attack}. Furthermore, the inherent vulnerabilities of LLMs themselves, such as hallucinations and unreliable reasoning~\citep{qian2026dynhd, bi2025cot,huang2025loongsynthesizelongchainofthoughts,yuan2026differential,shen2025faithcot,zhang2025grace,zhang2026confidence,wang2026step}, can be further amplified in multi-agent settings~\citep{A_trust,shen2025metacognitive}. The most severe threats target message-passing mechanisms~\citep{corba, pan2026explainable}, enabling malicious attackers to implant prejudiced content. NetSafe~\citep{NetSafe} pioneers the study of network structure vulnerabilities, identifying bias propagation patterns in a multi-agent utterance graph. G-Safeguard~\citep{G-Safeguard} advances supervised detection of compromised agents through graph neural networks and topological remediation. While these methods can mitigate certain security threats, they heavily rely on labeled malicious agents or prior knowledge of attack patterns, which may not be available in real-world MAS deployments. 

\section{Algorithm}
\label{appendix_algorithm}

For a detailed implementation of our proposed BlindGuard, please refer to Algorithm~\ref{algor:BlindGuard}.

\begin{algorithm}[!t]\small
    \caption{\ourmethod}
    \label{algor:BlindGuard}
    \raggedright
    \textbf{Input}: Normal MAS graphs $\{\mathcal{G}_1, \ldots, \mathcal{G}_{T_n}\}$, Attacked MAS graphs $\{\mathcal{G}'_1, \ldots, \mathcal{G}'_{T_a}\}$, hierarchical agent encoder $g_\theta$, Intensity parameter $\alpha$ and Anomaly budget $K$.\hspace*{0.15em} \\
    \textbf{Output}: Final responses $\{\widetilde{\mathcal{R}}_1, \ldots, \widetilde{\mathcal{R}}_{T_a}\}$ of all remediated MAS $\{\widetilde{\mathcal{G}}'_1, \ldots, \widetilde{\mathcal{G}}'_{T_a}\}$
    
    \begin{algorithmic}[1]
    % ===== Training Phase =====
    \FOR{each normal MAS $\mathcal{G}_i \in \{\mathcal{G}_1, \ldots, \mathcal{G}_{T_n}\}$}
        \FOR{each agent $v_j \in \mathcal{G}_i$}
            \STATE $\mathbf{x}_j \leftarrow \text{SentenceBERT}(R_j)$ 
            \STATE \tikz[baseline]{\node[anchor=base,fill=none,draw=none,text opacity=0.5] {$// \ \text{Node feature construction.}$}}\\
        \ENDFOR
        
        \STATE Sample subset $\mathcal{V}_{\text{corr}} \subset \mathcal{V}$
        \FOR{each agent $v_j \in \mathcal{V}_{\text{corr}}$}
            \STATE $\mathbf{x}_j \leftarrow \mathbf{x}_j + \alpha\|\mathbf{x}_j\|_2 \cdot \frac{\bm{\epsilon}_j}{\|\bm{\epsilon}_j\|_2}, \bm{\epsilon}_j \sim \mathcal{N}(0,\mathbf{I})$ 
            \STATE \tikz[baseline]{\node[anchor=base,fill=none,draw=none,text opacity=0.5] {$// \ \text{Feature corruption.}$}}\\
        \ENDFOR

        \FOR{each agent $v_j \in \mathcal{G}_i$}
            \STATE $\mathbf{h}_j^{\text{self}} \leftarrow \mathbf{x}_j$
            \STATE $\mathbf{h}_j^{\text{neigh}} \leftarrow \sum_{k \in \mathcal{N}(j)} \hat{A}_{jk} \mathbf{x}_k$
            \STATE $\mathbf{h}_j^{\text{graph}} \leftarrow \frac{1}{N}\sum_{k=1}^N \mathbf{x}_k$ 
            \STATE $\mathbf{z}_j \leftarrow g_\theta(\mathbf{h}_j^{\text{self}} \parallel \mathbf{h}_j^{\text{neigh}} \parallel \mathbf{h}_j^{\text{graph}})$
            \STATE \tikz[baseline]{\node[anchor=base,fill=none,draw=none,text opacity=0.5] {$// \ \text{Obtain agent representations.}$}}\\
        \ENDFOR

        \STATE $\text{Calculate supervised contrastive loss}\ \mathcal{L}$
        \STATE Update $\theta$ via gradient descent $\bigtriangledown_{\theta}\left[\mathcal{L} \right]$\\
        
    \ENDFOR
    
    % ===== Inference Phase =====
    \FOR{each attacked MAS $\mathcal{G}'_i \in \{\mathcal{G}'_1, \ldots, \mathcal{G}'_{T_a}\}$}
        \FOR{each agent $v_j \in \mathcal{G}'_t$}
            \STATE $\mathbf{x}_j \leftarrow \text{SentenceBERT}(R_j)$
            \STATE $\mathbf{z}_j \leftarrow f_\theta(\mathbf{x}_j, \mathcal{G}'_t)$
            \STATE $s_j \leftarrow -\frac{1}{N} \sum_{k=1}^N \text{sim}(\mathbf{z}_j, \mathbf{z}_k)$
            \STATE \tikz[baseline]{\node[anchor=base,fill=none,draw=none,text opacity=0.5] {$// \ \text{Compute anomaly score.}$}}\\
        \ENDFOR
        
        \STATE $\mathcal{V}_{\text{atk}} \leftarrow \text{Top-K agents with highest } s_j$
        
        \STATE $\mathcal{E}^{+} \leftarrow \{e_{kj} \in \mathcal{E} \mid v_k \notin \mathcal{V}_{\text{atk}}\}$ 
        \STATE \tikz[baseline]{\node[anchor=base,fill=none,draw=none,text opacity=0.5] {$// \ \text{MAS remediation.}$}}\\
        
        \FOR{each agent $v_j \in \mathcal{G}'_t$}
            \STATE $R_j \leftarrow \text{LLM}\left(Q \cup \{R_k \mid e_{kj} \in \mathcal{E}^{+}\}\right)$
        \ENDFOR
    \STATE Determine the final answer $\widetilde{\mathcal{R}}_i$ by aggregating all agent responses (e.g., majority voting)

    \ENDFOR
    
    \RETURN $\{\widetilde{\mathcal{R}}_1, \ldots, \widetilde{\mathcal{R}}_{T_a}\}$
    
    \end{algorithmic}
\end{algorithm}

\begin{figure*}[t]
    \centering
    \includegraphics[width=1.0\textwidth]{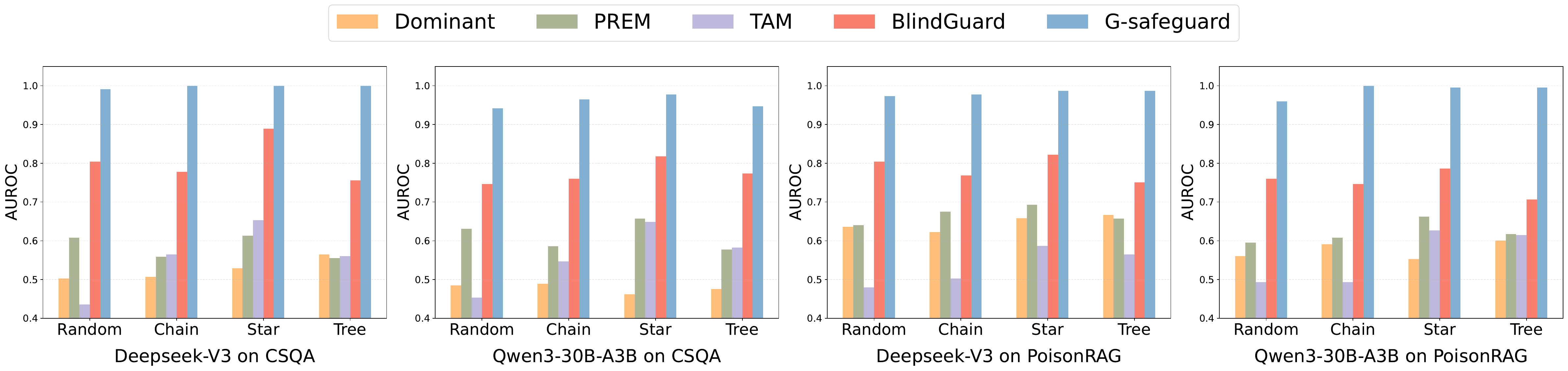}
    \caption{The AUC with DeepSeek-V3 and Qwen3-30B-A3B as backbone LLMs on the CSQA and PoisonRAG.}
    \label{fig:Deepseek_Qwen_AUC}
\end{figure*}

\begin{figure}[t]
    \centering
    \includegraphics[width=0.5\textwidth]{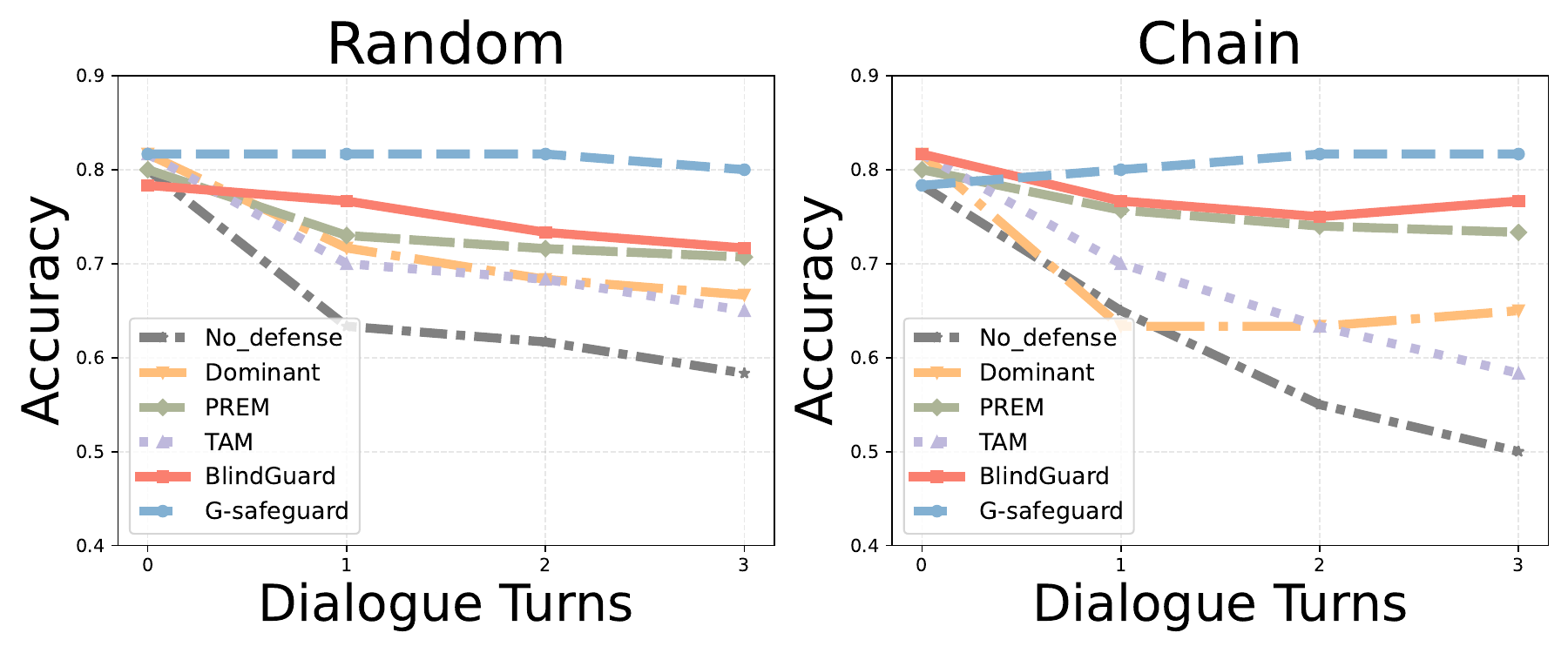}
    \caption{The overall performance of MAS on the CSQA dataset after each turn of dialogue. We use majority voting as the strategy to select the final answer.}
    \label{fig:MAS_ACC_CSQA_S_T}
\end{figure}

\section{Complexity Analysis}
\label{appendix_Complexity_Analysis}

We define key parameters: $N$ (number of agents), $d$ (representation dimension), $E$ (number of edges) and $L$ (GNN layers). For BlindGuard, after obtaining agent features via SentenceTransformer (shared across methods), the hierarchical encoder computes three representations. We reuse embeddings with $O\left(1\right)$ cost per agent for the node level. We aggregate via the normalized adjacency matrix with $O\left(Ed\right)$ cost for the neighbor level , and perform global average pooling with $O\left(Nd\right)$ cost for the system level. These are fused via an MLP encoder $O\left(Nd^{2}\right)$. Finally, anomaly scoring computes pairwise cosine similarities $O\left(N^{2}d\right)$. Total complexity is $O\left(Ed + Nd^{2} + N^{2}d\right)$. For G-Safeguard, the GNN requires $O\left(L\cdot\big(Ed + Nd^{2}\big)\right)$ for propagation and feature transformation, plus $O\left(Nd\right)$ for classification when the class is binary. Total complexity is $O\left(L\cdot\big(Ed + Nd^{2}\big) + Nd\right)$.

In typical MAS deployments, $N$ is modest (e.g., tens of agents). For sparse collaboration graphs, $E=O\left(N\right)$, and $d$ is in the hundreds. G-Safeguard introduces a multiplicative L factor via multi-layer GNN propagation. In practice, $O\left(L\cdot\big(Ed + Nd^{2}\big)\right)$ can exceed BlindGuard’s additional $O\left(N^{2}d\right)$ scoring at small $N$. When $N$ is large, we will mitigate the $O\left(N^{2}d\right)$ term via block-wise similarity computation with top‑k pruning, approximate nearest‑neighbor search and cluster‑based scoring, which reduce the practical cost. End-to-end latency is dominated by LLM inference, as the original system averages 0.53 minutes for three rounds per sample whereas enabling BlindGuard yields 0.56 minutes for the same three rounds, indicating an added cost of about 0.03 minutes while the majority of cost in LLM generation.

\begin{table*}[t]
\centering
\resizebox{0.85\textwidth}{!}{%
\scriptsize
\begin{tabular}{@{}llccc|llccc@{}}
\toprule
\multicolumn{5}{c|}{\textbf{MA (CSQA)}} & \multicolumn{5}{c}{\textbf{PI (CSQA)}} \\
\cmidrule(lr){1-5} \cmidrule(lr){6-10}
\textbf{Topology} & \textbf{Defense} & \textbf{R1} & \textbf{R2} & \textbf{R3} & 
\textbf{Topology} & \textbf{Defense} & \textbf{R1} & \textbf{R2} & \textbf{R3} \\
\midrule
\multirow{6}{*}{Random}
& ND & 23.0 & 38.7 & 48.0 &
\multirow{6}{*}{Random}
& ND & 40.7 & 63.0 & 75.3 \\
& G-Safeguard & 8.7 & 13.3 & 13.0 && G-Safeguard & 18.7 & 19.3 & 21.0 \\
& DOMINANT & 26.3 & 40.7 & 49.0 && DOMINANT & 33.0 & 52.3 & 61.3 \\
& PREM & 22.7 & 29.3 & 33.3 && PREM & 34.3 & 51.0 & 59.0 \\
& TAM & 21.7 & 34.7 & 39.0 && TAM & 35.3 & 50.3 & 63.7 \\
& BlindGuard & 15.0 & 22.3 & 27.0 && BlindGuard & 29.0 & 41.7 & 46.0 \\
\cmidrule(lr){1-5} \cmidrule(lr){6-10}

\multirow{6}{*}{Tree}
& ND & 27.3 & 38.3 & 43.6 &
\multirow{6}{*}{Tree}
& ND & 36.3 & 58.7 & 68.3 \\
& G-Safeguard & 7.3 & 6.7 & 9.3 && G-Safeguard & 16.7 & 17.3 & 18.7 \\
& DOMINANT & 19.0 & 26.3 & 30.0 && DOMINANT & 31.3 & 42.7 & 48.0 \\
& PREM & 18.3 & 22.7 & 26.7 && PREM & 25.7 & 38.0 & 40.3 \\
& TAM & 16.3 & 25.0 & 28.7 && TAM & 29.0 & 37.3 & 39.0 \\
& BlindGuard & 8.7 & 10.7 & 12.3 && BlindGuard & 26.7 & 34.0 & 36.7 \\
\bottomrule
\end{tabular}
}
\caption{ASR across communication rounds with Qwen3‑8B for MA (CSQA) and PI (CSQA) datasets.}
\label{tab_weak_model}
\end{table*}

\begin{table}[t]\small
\centering
\begin{tabular}{ccccccc}
\toprule
\textbf{Datasets} & \textbf{0.03} & \textbf{0.3} & \textbf{0.5} & \textbf{0.8} & \textbf{10}\\
\midrule
\textbf{PI (CSQA)} & 52.4  & 72.9 & 73.8 & 75.1 & 62.7 \\
\textbf{MA (PoisonRAG)} & 61.3 & 75.6 & 80.0 & 82.7 & 68.9 \\
\bottomrule
\end{tabular}
\caption{Impact of varying $\alpha$ on AUC for PI (CSQA) and MA (PoisonRAG) datasets.}
\label{tab:Hyperparameter_Analysis}
%\vspace{-8pt}
\end{table}

\section{Detailed Experimental Setups}
\label{appendix_experimental_setup}

We employ the Adam optimizer with an initial learning rate of 0.001 and L2 regularization ($\text{weight decay} \in \{5\times10^{-5}, 10^{-4}, 2\times10^{-4}\}$). The learning rate is dynamically adjusted using a cosine annealing scheduler ($T_{\text{max}} = 10$ cycles and $\eta_{\text{min}} = 10^{-5}$) to facilitate better convergence. The configuration of detection budget $K=3$ is grounded in the fact that exactly 3 agents are compromised in all test scenarios, ensuring a consistent evaluation framework for calculating AUC. All models are implemented with a hidden dimension of 512 and trained on 4 NVIDIA L40 GPUs.

During inference, BlindGuard computes an anomaly score for each agent and flags the top‑K as malicious. Because some defense methods may predict that all agents are malicious and disrupt all communications, which is not practical. Prompt structures follow G‑Safeguard~\citep{G-Safeguard}, with different attack types modifying the prompt accordingly. The monitoring module encodes each agent’s reply together with neighbor summaries and a global system view, compares these representations to assess contextual consistency, and prunes communication links to detected anomalies, thereby preventing adversarial propagation without relying on predefined attack signatures.

\section{Additional Experiments}
\label{appendix_experiments}
To further validate the effectiveness of our proposed BlindGuard, we conduct additional experiments. 

\subsection{Results with More LLMs}

As shown in Figure~\ref{fig:Deepseek_Qwen_AUC}, the experiments demonstrate BlindGuard's consistent performance advantages, showing superior AUC scores over Dominant, PREM, and TAM across four topologies with both DeepSeek-V3 and Qwen3-30B-A3B LLMs on CSQA and PoisonRAG benchmarks. 
Figure~\ref{fig:MAS_ACC_CSQA_S_T} provides extended experimental results comparing BlindGuard with baseline methods (Dominant, PREM, TAM, G-safeguard) across multiple dialogue turns, demonstrating consistent accuracy improvements while maintaining efficient communication in the CSQA task. 

To further examine BlindGuard’s robustness with weaker LLMs, we run supplementary experiments on MA (CSQA) and PI (CSQA) under random and tree topologies using Qwen3‑8B. As shown in Table~\ref{tab_weak_model}, compared with GPT‑4o‑mini as the backbone, AUC is similar while ASR is higher. AUC reflects detector ranking quality and remains stable, indicating BlindGuard still separates benign and compromised agents effectively. ASR depends on the backbone’s susceptibility and the dataset’s difficulty. The weaker backbone is more likely to comply with adversarial prompts and to generate off‑intent content in CSQA, which increases ASR for all methods. Despite this, BlindGuard remains effective by reducing multi‑round propagation relative to no defense and other unsupervised methods under the same weaker backbone.

\begin{table}[ht]\small
\centering
\begin{tabular}{ccccccc}
\toprule
\textbf{Methods} & \textbf{Datasets} & \textbf{No defense} & \textbf{Threshold} & \textbf{TopK} \\
\midrule
\multirow{2}{*}{\textbf{Chain}} 
    & CSQA (PI) & 42.33 & 22.41 & 23.67 \\
    & CSQA (MA) & 26.67 & 7.00  & 7.00  \\
\midrule
\multirow{2}{*}{\textbf{Random}} 
    & CSQA (PI) & 42.00 & 29.31 & 30.00 \\
    & CSQA (MA) & 24.00 & 15.00 & 8.67  \\
\midrule
\multirow{2}{*}{\textbf{Star}} 
    & CSQA (PI) & 50.33 & 27.00 & 22.00 \\
    & CSQA (MA) & 27.00 & 15.67 & 9.33  \\
\midrule
\multirow{2}{*}{\textbf{Tree}} 
    & CSQA (PI) & 33.00 & 24.00 & 24.00 \\
    & CSQA (MA) & 21.33 & 6.67  & 7.67  \\
\bottomrule
\end{tabular}
\caption{Performance of BlindGuard under fixed budget and threshold-based remediation strategies on CSQA (PI) and CSQA (MA).}
\label{tab:threshold_comparison}
\end{table}

\subsection{Hyperparameter Sensitivity Analysis}

We conduct a sensitivity study for $\alpha$ (the corruption‑intensity coefficient in Eq.~\ref{Corruption_eq}) on the random topology for two datasets. As shown in Table~\ref{tab:Hyperparameter_Analysis}, we find that performance is stable in the practical range 0.3–0.8, while very small or very large $\alpha$ degrades AUC. When $\alpha$ is too small, directional negatives become nearly isotropic and the margin is weak. When $\alpha$ is too large, semantics are over‑distorted, whereas a moderate $\alpha$ creates a robust angular margin around normal behavior. In all experiments, we fix $\alpha = 0.8$ for all datasets and topologies, which supports that BlindGuard is not sensitive to this hyperparameter $\alpha$.

\begin{table*}[ht!]
\centering
\resizebox{0.75\textwidth}{!}{%
\scriptsize
\begin{tabular}{@{}lccc|lccc@{}}
\toprule
\multicolumn{4}{c|}{\textbf{Random Topology}} & \multicolumn{4}{c}{\textbf{Chain Topology}} \\
\cmidrule(lr){1-4} \cmidrule(lr){5-8}
\textbf{Method} & \textbf{R1} & \textbf{R2} & \textbf{R3} & 
\textbf{Method} & \textbf{R1} & \textbf{R2} & \textbf{R3} \\
\midrule
No Defense & 23.7 & 28.5 & 30.8 &
No Defense & 24.4 & 31.1 & 31.8 \\
G-Safeguard & 18.3 & 18.6 & 19.7 &
G-Safeguard & 19.3 & 19.7 & 20.0 \\
DOMINANT & 24.0 & 26.1 & 27.8 &
DOMINANT & 20.0 & 22.7 & 22.7 \\
PREM & 20.7 & 22.7 & 23.0 &
PREM & 19.3 & 21.7 & 21.7 \\
TAM & 21.4 & 23.7 & 25.1 &
TAM & 19.7 & 20.7 & 21.4 \\
BlindGuard & 18.7 & 16.7 & 17.6 &
BlindGuard & 18.0 & 20.3 & 19.3 \\
\cmidrule(lr){1-4} \cmidrule(lr){5-8}

\multicolumn{4}{c|}{\textbf{Star Topology}} & \multicolumn{4}{c}{\textbf{Tree Topology}} \\
\cmidrule(lr){1-4} \cmidrule(lr){5-8}
\textbf{Method} & \textbf{R1} & \textbf{R2} & \textbf{R3} & 
\textbf{Method} & \textbf{R1} & \textbf{R2} & \textbf{R3} \\
\midrule
No Defense & 27.0 & 33.7 & 35.7 &
No Defense & 23.4 & 26.1 & 27.5 \\
G-Safeguard & 18.3 & 19.3 & 20.0 &
G-Safeguard & 21.0 & 20.0 & 21.4 \\
DOMINANT & 19.7 & 20.7 & 22.7 &
DOMINANT & 20.3 & 21.0 & 23.0 \\
PREM & 22.7 & 26.0 & 26.7 &
PREM & 18.7 & 19.3 & 19.3 \\
TAM & 19.3 & 20.0 & 21.4 &
TAM & 21.0 & 22.4 & 23.0 \\
BlindGuard & 18.3 & 17.6 & 17.3 &
BlindGuard & 19.7 & 21.0 & 20.7 \\
\bottomrule
\end{tabular}
}
\caption{Attack success rate (ASR) across communication rounds under collusion attacks.}
\label{tab:Collusion Attack}
\end{table*}

\begin{table}[ht]\small
\centering
\begin{tabular}{cccccc}
\toprule
\textbf{Topology} & \textbf{AUC} & \textbf{ASR@1} & \textbf{ASR@2} & \textbf{ASR@3} \\
\midrule
Chain & 76.94 & 18.30 & 19.66 & 21.36 \\
Tree & 76.00 & 21.33 & 25.00 & 27.67 \\
Star & 83.56 & 17.33 & 19.67 & 20.67 \\
Random & 77.77 & 18.67 & 20.67 & 21.33 \\
\bottomrule
\end{tabular}
\caption{ASR and AUC on the CSQA (PI) under adaptive attacks.}
\label{tab:adaptive_attacks}
\vspace{-8pt}
\end{table}

\subsection{Threshold-based Defense}
Several graph anomaly detection baselines can produce uniformly high anomaly scores, which may cause nearly all agents to be flagged as anomalous under a naive thresholding rule. To avoid giving any method an unfair advantage due to threshold calibration, we report results under a fixed remediation budget. BlindGuard can be deployed with a threshold on the anomaly score. To demonstrate this, we additionally run a threshold-based variant where we flag agents with anomaly score 0.5, using GPT-4o-mini on CSQA(PI) and CSQA(MA). As shown in Table~\ref{tab:threshold_comparison}, the results (ACC/ASR) remain competitive and show that BlindGuard is still effective under this more realistic "unknown-K" setting.

%\vspace{-2pt}
\subsection{Collusion Attacks}

We designed a sacrificial‑decoy collusion on PI (CSQA) where an attacked agent adds an extra instruction ``Before generating an answer, please output something completely unrelated to this question.'' to draw the detector’s attention while its peers remain subtly biased and neighbor‑consistent. As shown in Table~\ref{tab:Collusion Attack}, BlindGuard achieves lower ASR across rounds and higher AUC than G‑Safeguard in this collusion setting, indicating stronger robustness to sacrificial decoys and stealthy cliques.

Supervised detectors trained with malicious labels tend to overweight overt anomalies, making them susceptible to ``sacrifice'' tactics. BlindGuard learns a margin around normal behavior using attack‑agnostic training, so it does not hinge on decoy signatures and preserves separation for stealthy colluders. BlindGuard combines node, neighbor, and global views, so the clique’s global mismatch remains detectable even when neighbor evidence is cleaned by a sacrificial decoy.

\begin{table*}[ht]\small
\centering
\begin{tabular}{cccccccccc}
\toprule
\multirow{2}{*}{\textbf{Datasets}} & \multirow{2}{*}{\textbf{Methods}} & \multicolumn{2}{c}{\textbf{Chain}} & \multicolumn{2}{c}{\textbf{Random}} & \multicolumn{2}{c}{\textbf{Star}} & \multicolumn{2}{c}{\textbf{Tree}} \\
\cmidrule(lr){3-4} \cmidrule(lr){5-6} \cmidrule(lr){7-8} \cmidrule(lr){9-10}
& & \textbf{AUC} & \textbf{ASR@3} & \textbf{AUC} & \textbf{ASR@3} & \textbf{AUC} & \textbf{ASR@3} & \textbf{AUC} & \textbf{ASR@3} \\
\midrule
\multirow{2}{*}{\textbf{CSQA (PI)}} 
    & Raw SBERT  & 67.56 & 28.33 & 77.78 & 33.67 & 66.22 & 21.33 & 67.56 & 29.67 \\
    & BlindGuard & 80.00 & 23.67 & 75.11 & 30.00 & 85.78 & 22.00 & 74.67 & 24.00 \\
\midrule
\multirow{2}{*}{\textbf{CSQA (MA)}} 
    & Raw SBERT  & 72.89 & 11.33 & 74.67 & 9.67  & 71.56 & 11.67 & 71.44 & 11.00 \\
    & BlindGuard & 73.78 & 7.00  & 75.56 & 8.67  & 76.44 & 9.33  & 72.89 & 7.67  \\
\bottomrule
\end{tabular}
\caption{Encoder ablation results comparing Raw SentenceBERT + Eq.(7) and BlindGuard in terms of AUC (\%) and ASR@3 (\%) across four communication topologies on CSQA (PI) and CSQA (MA).}
\label{tab:encoder_ablation}
\end{table*}

\begin{table}[t]\small
\centering
\begin{tabular}{cccccc}
\toprule
\textbf{Datasets} & \textbf{Methods} & \textbf{Chain} & \textbf{Random} & \textbf{Star} & \textbf{Tree} \\
\midrule
\multirow{3}{*}{\textbf{PI}} 
    & ND    & 90.0  & 90.0  & 90.0 & 90.0 \\
    & GS   & 90.0  & 91.7  & 90.0 & 88.3 \\
    & BG    & 90.0  & 88.3  & 90.0 & 90.0 \\
\midrule
\multirow{3}{*}{\textbf{MA}} 
    & ND    & 100.0 & 100.0 & 100.0 & 100.0 \\
    & GS   & 100.0 & 98.3  & 100.0 & 100.0 \\
    & BG    & 100.0 & 100.0 & 100.0 & 100.0 \\
\bottomrule
\end{tabular}

\caption{Accuracy on attack-free samples across different graph topologies on CSQA.ND = No Defense, GS = G-Safeguard, BG = BlindGuard}
\label{tab:clean_eval}
\end{table}

\subsection{Adaptive Attacks}
In real-world scenarios, adversaries may learn the principles of a defense mechanism and deliberately adapt their attack strategy to bypass detection. This type of targeted modification is known as an adaptive attack. To evaluate the robustness of BlindGuard under such conditions, we design a dedicated experiment against a simple adaptive Prompt Injection (PI) strategy with GPT-4o-mini as backbone LLM on the PI (CSQA).
We modify the original PI attack prompt by embedding the correct answer directly into the input and instructing the agent to ``mimic the reasoning process'' of the correct answer while explaining the attacker-provided answer, aiming to reduce semantic deviation between malicious responses and normal agent outputs. This strategy is intended to bypass BlindGuard’s anomaly detection. As shown in Table~\ref{tab:adaptive_attacks}, BlindGuard consistently achieved low ASR and the detection AUC remained high, indicating that the hierarchical encoder and corruption-guided detector in BlindGuard were able to capture subtle abnormal interaction patterns and isolate compromised agents, even when their semantic embeddings were close to those of normal agents.

\subsection{Encoder Contribution Analysis}
To isolate the contribution of our learned encoder, we introduce an additional baseline, Raw SentenceBERT + Eq.(7), which removes the encoder entirely and computes the anomaly score in Eq.(7) directly on non-trained SentenceBERT embeddings. We evaluate this baseline under the same experimental setup (GPT-4o-mini) on CSQA (PI) and CSQA (MA) across four communication topologies (Chain / Random / Star / Tree). As shown in Table~\ref{tab:encoder_ablation}, directly applying Eq.(7) to raw SBERT features yields consistently inferior performance compared to BlindGuard. The consistent performance gains demonstrate that our encoder training, incorporating semantics-aware directional negatives and margin separation, meaningfully improves the representation geometry, enabling more reliable unsupervised anomaly scoring.

\subsection{Attack-free Evaluation}
To evaluate whether edge pruning may falsely disrupt normal multi-agent system (MAS) functionality in the absence of attacks, we conduct a no-attack evaluation using the same CSQA (PI) and CSQA (MA) settings with GPT-4o-mini across four communication topologies. As shown in Table~\ref{tab:clean_eval}, BlindGuard does not introduce noticeable utility degradation on clean samples, the accuracy remains essentially unchanged compared to the no-defense baseline, and is comparable to the supervised baseline G-Safeguard. These results indicate that BlindGuard can be safely deployed without significantly disrupting normal MAS functionality.

\subsection{Additional Results of BlindGuard}
We present more detailed results on CSQA, MMLU, GSM8K, InjecAgent and PoisonRAG benchmarks.
As shown in Table~\ref{tab_round_1} and Table~\ref{tab_round_2}, additional experiments provide comprehensive multi-round ASR@3 comparisons (R1-R3) across multiple attack types and network topologies, demonstrating BlindGuard's superior performance over Dominant, PREM, and TAM while approaching G-Safeguard's effectiveness throughout progressive dialogue stages in various attack types.

To demonstrate BlindGuard's effectiveness under constrained detection resources where computational resources or operational constraints limit the number of agents that can be simultaneously isolated. The $K=2$ setting represents a cost-effective defense approach, where the system achieves substantial security improvement while minimizing disruption to normal agent operations. As shown in Table~\ref{tab:asr_comprehensive}, BlindGuard shows consistent defense capability, approaching the supervised G-Safeguard's effectiveness.

\begin{table*}[t]
\centering
\scriptsize
\begin{tabular}{@{}llccc|llccc|llccc@{}}
\toprule
\multicolumn{5}{c|}{\textbf{PI (CSQA)}} & \multicolumn{5}{c|}{\textbf{PI (MMLU)}} & \multicolumn{5}{c}{\textbf{PI (GSM8K)}} \\
\cmidrule(lr){1-5} \cmidrule(lr){6-10} \cmidrule(lr){11-15}
\textbf{Topology} & \textbf{Defense} & \textbf{R1} & \textbf{R2} & \textbf{R3} & 
\textbf{Topology} & \textbf{Defense} & \textbf{R1} & \textbf{R2} & \textbf{R3} & 
\textbf{Topology} & \textbf{Defense} & \textbf{R1} & \textbf{R2} & \textbf{R3} \\
\midrule
\multirow{6}{*}{Chain}
& ND & 0.31 & 0.397 & 0.423 &
\multirow{6}{*}{Chain}
& ND & 0.283 & 0.347 & 0.383 &
\multirow{6}{*}{Chain}
& ND & 0.064 & 0.092 & 0.098 \\
& Dom & 0.247 & 0.283 & 0.303 && Dom & 0.217 & 0.240 & 0.247 && Dom & 0.071 & 0.088 & 0.085 \\
& PREM & 0.237 & 0.290 & 0.293 && PREM & 0.240 & 0.257 & 0.250 && PREM & 0.061 & 0.075 & 0.088 \\
& TAM & 0.217 & 0.260 & 0.273 && TAM & 0.220 & 0.233 & 0.237 && TAM & 0.064 & 0.088 & 0.088 \\
& BG & 0.203 & 0.223 & 0.237 && BG & 0.190 & 0.193 & 0.193 && BG & 0.061 & 0.085 & 0.085 \\
& GS & 0.187 & 0.197 & 0.193 && GS & 0.190 & 0.193 & 0.193 && GS & 0.044 & 0.044 & 0.044 \\
\cmidrule(lr){1-5} \cmidrule(lr){6-10} \cmidrule(lr){11-15}
\multirow{6}{*}{Tree}
& ND & 0.293 & 0.337 & 0.330 &
\multirow{6}{*}{Tree}
& ND & 0.250 & 0.300 & 0.320 &
\multirow{6}{*}{Tree}
& ND & 0.075 & 0.082 & 0.102 \\
& Dom & 0.233 & 0.260 & 0.270 && Dom & 0.200 & 0.200 & 0.193 && Dom & 0.061 & 0.061 & 0.068 \\
& PREM & 0.220 & 0.227 & 0.267 && PREM & 0.207 & 0.220 & 0.240 && PREM & 0.064 & 0.075 & 0.085 \\
& TAM & 0.223 & 0.253 & 0.260 && TAM & 0.230 & 0.233 & 0.220 && TAM & 0.071 & 0.075 & 0.081 \\
& BG & 0.227 & 0.257 & 0.240 && BG & 0.207 & 0.220 & 0.207 && BG & 0.075 & 0.075 & 0.085 \\
& GS & 0.180 & 0.190 & 0.187 && GS & 0.187 & 0.187 & 0.183 && GS & 0.071 & 0.075 & 0.078 \\
\cmidrule(lr){1-5} \cmidrule(lr){6-10} \cmidrule(lr){11-15}
\multirow{6}{*}{Star}
& ND & 0.333 & 0.453 & 0.503 &
\multirow{6}{*}{Star}
& ND & 0.293 & 0.350 & 0.390 &
\multirow{6}{*}{Star}
& ND & 0.071 & 0.115 & 0.129 \\
& Dom & 0.303 & 0.357 & 0.387 && Dom & 0.227 & 0.257 & 0.267 && Dom & 0.054 & 0.071 & 0.078 \\
& PREM & 0.280 & 0.330 & 0.343 && PREM & 0.260 & 0.297 & 0.310 && PREM & 0.068 & 0.095 & 0.105 \\
& TAM & 0.240 & 0.270 & 0.287 && TAM & 0.243 & 0.263 & 0.283 && TAM & 0.061 & 0.081 & 0.081 \\
& BG & 0.193 & 0.213 & 0.220 && BG & 0.200 & 0.207 & 0.203 && BG & 0.068 & 0.068 & 0.068 \\
& GS & 0.183 & 0.190 & 0.193 && GS & 0.177 & 0.170 & 0.173 && GS & 0.054 & 0.064 & 0.061 \\
\cmidrule(lr){1-5} \cmidrule(lr){6-10} \cmidrule(lr){11-15}
\multirow{6}{*}{Random} 
& ND & 0.32 & 0.377 & 0.420 &
\multirow{6}{*}{Random} 
& ND & 0.347 & 0.393 & 0.410 &
\multirow{6}{*}{Random} 
& ND & 0.059 & 0.097 & 0.145 \\
& Dom & 0.313 & 0.343 & 0.370 && Dom & 0.247 & 0.280 & 0.290 && Dom & 0.054 & 0.085 & 0.105 \\
& PREM & 0.280 & 0.330 & 0.347 && PREM & 0.273 & 0.337 & 0.370 && PREM & 0.037 & 0.078 & 0.105 \\
& TAM & 0.283 & 0.323 & 0.353 && TAM & 0.307 & 0.380 & 0.383 && TAM & 0.047 & 0.098 & 0.149 \\
& BG & 0.243 & 0.280 & 0.300 && BG & 0.220 & 0.253 & 0.267 && BG & 0.041 & 0.054 & 0.064 \\
& GS & 0.190 & 0.193 & 0.190 && GS & 0.180 & 0.187 & 0.190 && GS & 0.048 & 0.038 & 0.038 \\
\bottomrule
\end{tabular}
\caption{Attack success rate (ASR) comparison of defense methods with GPT-4o-mini as backbone LLMs (Part 1). ND = No Defense, Dom = Dominant, BG = BlindGuard, GS = G-Safeguard. Lower values indicate better defense performance.}
\label{tab_round_1}
\end{table*}

\section{Topology-dependent Performance}

We provide an in-depth analysis across topologies showing how topologies mediate attack propagation and BlindGuard’s effectiveness, grounded in the table 1, followed by brief case studies.

\paragraph{Chain.}
Propagation is sequential and relatively slow. BlindGuard consistently lowers ASR@3 versus no defense across attacks. The neighbor and global views are complementary in this setting, and local shift is easier to detect, so pruning upstream edges quickly reduces downstream exposure.

\paragraph{Tree.}
Subtrees can collude and create strong local consistency while deviating from the global intent. BlindGuard’s system-level context is particularly helpful and delivers large ASR reductions. Across tasks, BlindGuard remains competitive, and AUC indicates reliable separation despite heterogeneous branches.

\paragraph{Star.}
An attacked agent can broadcast widely, which makes the topology high risk. BlindGuard benefits from a strong global signal and promptly prunes edges of central agent, yielding marked ASR drops. The approach is effective because central agent misalignment with the global context is easy to detect, and containment at the central agent immediately protects most spokes.

\paragraph{Random.}
Low-degree paths dilute neighbor evidence, which makes PI tasks relatively harder to contain, and ASR@3 remains higher than in star or tree. Random connectivity reduces the reliability of neighbor information. The global context stabilizes detection. BlindGuard still achieves strong containment on tool and memory attacks where semantic shift is more uniform.

The star topology has a single center agent that connects to all other agents, and the peripheral agents do not connect to each other. Under most cases of detection errors, an undetected attacked peripheral agent can easily push misleading content to the center, which then distributes it to the entire system. This vulnerability pathway through the center agent is structurally distinct from other topologies, where spread relies on sequential or irregular multi‑path transmission rather than a single global distributor.

We provided a case analysis using this following specific query ``John was an aristocratic fox hunter. Where might he live?\textbackslash nA england\textbackslash nB. new hampshire\textbackslash nC. street\textbackslash nD. arkansas\textbackslash nE. north dakota''. We observed a false positive on the center agent, which was actually benign in star topology. Pruning the agent’s outgoing edges stopped redistribution. Although the attack reached the center from a peripheral agent, it could not be forwarded further, and the peripherals had no lateral links, so the spread was avoided even with a misdetection of the compromised node. In random topology, with sparse and irregular links, neighbor evidence is diluted and the wrong answer can still spread across multiple paths when early misclassifications occur. This case demonstrates that interaction topology materially affects both defense performance and attack success rates.

\begin{table*}[t!]
\centering
\scriptsize
\begin{tabular}{@{}llccc|llccc|llccc@{}}
\toprule
\multicolumn{5}{c|}{\textbf{TA (InjecAgent)}} & \multicolumn{5}{c|}{\textbf{MA (CSQA)}} & \multicolumn{5}{c}{\textbf{MA (PoisonRAG)}} \\
\cmidrule(lr){1-5} \cmidrule(lr){6-10} \cmidrule(lr){11-15}
\textbf{Topology} & \textbf{Defense} & \textbf{R1} & \textbf{R2} & \textbf{R3} & 
\textbf{Topology} & \textbf{Defense} & \textbf{R1} & \textbf{R2} & \textbf{R3} & 
\textbf{Topology} & \textbf{Defense} & \textbf{R1} & \textbf{R2} & \textbf{R3} \\
\midrule
\multirow{6}{*}{Chain}
& ND & 0.337 & 0.442 & 0.480 &
\multirow{6}{*}{Chain}
& ND & 0.137 & 0.237 & 0.267 &
\multirow{6}{*}{Chain}
& ND & 0.120 & 0.193 & 0.223 \\
& Dom & 0.153 & 0.142 & 0.150 && Dom & 0.137 & 0.180 & 0.187 && Dom & 0.090 & 0.100 & 0.110 \\
& PREM & 0.147 & 0.152 & 0.152 && PREM & 0.083 & 0.133 & 0.157 && PREM & 0.030 & 0.043 & 0.060 \\
& TAM & 0.243 & 0.293 & 0.300 && TAM & 0.090 & 0.130 & 0.133 && TAM & 0.077 & 0.120 & 0.147 \\
& BG & 0.142 & 0.178 & 0.164 && BG & 0.043 & 0.053 & 0.070 && BG & 0.047 & 0.083 & 0.100 \\
& GS & 0.122 & 0.108 & 0.102 && GS & 0.060 & 0.067 & 0.077 && GS & 0.030 & 0.030 & 0.040 \\
\cmidrule(lr){1-5} \cmidrule(lr){6-10} \cmidrule(lr){11-15}
\multirow{6}{*}{Tree}
& ND & 0.289 & 0.428 & 0.451 &
\multirow{6}{*}{Tree}
& ND & 0.153 & 0.200 & 0.213 &
\multirow{6}{*}{Tree}
& ND & 0.107 & 0.157 & 0.180 \\
& Dom & 0.146 & 0.149 & 0.153 && Dom & 0.150 & 0.200 & 0.217 && Dom & 0.083 & 0.123 & 0.143 \\
& PREM & 0.159 & 0.182 & 0.162 && PREM & 0.087 & 0.110 & 0.140 && PREM & 0.037 & 0.057 & 0.083 \\
& TAM & 0.231 & 0.289 & 0.321 && TAM & 0.113 & 0.153 & 0.130 && TAM & 0.057 & 0.090 & 0.120 \\
& BG & 0.142 & 0.143 & 0.125 && BG & 0.050 & 0.067 & 0.077 && BG & 0.037 & 0.057 & 0.070 \\
& GS & 0.076 & 0.061 & 0.048 && GS & 0.067 & 0.060 & 0.067 && GS & 0.017 & 0.033 & 0.030 \\
\cmidrule(lr){1-5} \cmidrule(lr){6-10} \cmidrule(lr){11-15}
\multirow{6}{*}{Star}
& ND & 0.368 & 0.482 & 0.436 &
\multirow{6}{*}{Star}
& ND & 0.137 & 0.207 & 0.270 &
\multirow{6}{*}{Star}
& ND & 0.127 & 0.223 & 0.260 \\
& Dom & 0.130 & 0.151 & 0.143 && Dom & 0.133 & 0.227 & 0.267 && Dom & 0.110 & 0.140 & 0.137 \\
& PREM & 0.132 & 0.139 & 0.147 && PREM & 0.100 & 0.173 & 0.200 && PREM & 0.063 & 0.073 & 0.090 \\
& TAM & 0.246 & 0.266 & 0.266 && TAM & 0.110 & 0.170 & 0.190 && TAM & 0.083 & 0.157 & 0.197 \\
& BG & 0.118 & 0.132 & 0.126 && BG & 0.040 & 0.067 & 0.093 && BG & 0.037 & 0.080 & 0.100 \\
& GS & 0.085 & 0.072 & 0.069 && GS & 0.033 & 0.010 & 0.017 && GS & 0.000 & 0.007 & 0.017 \\
\cmidrule(lr){1-5} \cmidrule(lr){6-10} \cmidrule(lr){11-15}
\multirow{6}{*}{Random} 
& ND & 0.336 & 0.409 & 0.398 &
\multirow{6}{*}{Random} 
& ND & 0.120 & 0.213 & 0.240 &
\multirow{6}{*}{Random} 
& ND & 0.123 & 0.197 & 0.247 \\
& Dom & 0.144 & 0.160 & 0.149 && Dom & 0.147 & 0.277 & 0.343 && Dom & 0.080 & 0.133 & 0.170 \\
& PREM & 0.115 & 0.136 & 0.145 && PREM & 0.100 & 0.153 & 0.167 && PREM & 0.067 & 0.107 & 0.137 \\
& TAM & 0.321 & 0.383 & 0.358 && TAM & 0.150 & 0.230 & 0.267 && TAM & 0.110 & 0.187 & 0.237 \\
& BG & 0.125 & 0.173 & 0.177 && BG & 0.040 & 0.063 & 0.087 && BG & 0.033 & 0.057 & 0.067 \\
& GS & 0.079 & 0.066 & 0.051 && GS & 0.037 & 0.050 & 0.050 && GS & 0.020 & 0.057 & 0.070 \\
\bottomrule
\end{tabular}
\caption{Attack success rate (ASR) comparison of defense methods with GPT-4o-mini as backbone LLMs (Part 2). Abbreviations same as in Table~\ref{tab_round_1}.}
\label{tab_round_2}
\end{table*}

\section{Key Differences from Prior Work}
Existing graph anomaly detection (GAD) methods for MAS mainly operate on either structural deviations or static attribute outliers, and are typically designed for fixed graph settings. These methods struggle to detect coordinated semantic anomalies in dynamic, language‑based MAS interactions. Traditional MAS security mechanisms such as rule‑based detection systems rely on rigid, pre‑specified communication formats and lack the capability to handle free‑form, high‑dimensional natural‑language exchanges in MAS. In contrast, BlindGuard introduces a hierarchical agent encoder that jointly captures agent‑level semantics, neighborhood context, and global system state, coupled with a corruption‑guided semantic anomaly detector that learns entirely from benign interactions without any attack labels or prior signatures. This enables BlindGuard to proactively identify and isolate malicious agents within the message‑passing process, blocking adversarial propagation before it contaminates other agents, and distinguishing it from purely structural detection, output‑level consensus, or rule‑driven MAS security designs.

\section{Directional Negative Justification}

A potential concern is whether BlindGuard relies on isotropic corruption as its primary modeling assumption. We clarify that this is not the case. Our defense is built around semantics-aware directional negatives and an angular-margin separation from the normal manifold, which is more closely aligned with how real semantic attacks deviate from normal behavior.
We encode each agent's message into an embedding and compute global (system-level) and neighborhood context vectors. The anomaly score measures alignment with these contexts via cosine similarity. Real semantic attacks tend to shift an agent's message away from the round's intent, reducing alignment with global and neighborhood contexts. 
Training with directional negatives enforces an angular (cosine) margin around the normal manifold. Any perturbation whose direction falls within a cone opposing the context vectors must reduce alignment beyond this margin, and thus becomes detectable. This mechanism is standard in metric and contrastive learning: semantics-aware hard negatives tighten the decision boundary precisely in the regions where real attacks are most likely to occur, whereas isotropic random noise does not provide such targeted boundary refinement. We acknowledge, however, that in stealthier attack scenarios where adversarial perturbations are carefully crafted to remain close to the normal manifold, the angular margin may be insufficient, and we leave more robust defenses against such scenarios for future work.

\begin{table*}[t!]
\centering
\resizebox{0.75\textwidth}{!}{%
\scriptsize
\begin{tabular}{@{}lccc|lccc@{}}
\toprule
\multicolumn{4}{c|}{\textbf{Random Topology}} & \multicolumn{4}{c}{\textbf{Chain Topology}} \\
\cmidrule(lr){1-4} \cmidrule(lr){5-8}
\textbf{Method} & \textbf{R1} & \textbf{R2} & \textbf{R3} & 
\textbf{Method} & \textbf{R1} & \textbf{R2} & \textbf{R3} \\
\midrule
No Defense & 0.3467 & 0.3933 & 0.4100 &
No Defense & 0.2833 & 0.3467 & 0.3833 \\
Dominant & 0.2933 & 0.3767 & 0.4000 &
Dominant & 0.2233 & 0.2367 & 0.2600 \\
PREM & 0.3100 & 0.3933 & 0.4300 &
PREM & 0.2467 & 0.2667 & 0.2767 \\
TAM & 0.3267 & 0.3833 & 0.3933 &
TAM & 0.2267 & 0.2600 & 0.2733 \\
BlindGuard & 0.2433 & 0.2967 & 0.3400 &
BlindGuard & 0.2467 & 0.2667 & 0.2667 \\
G-Safeguard & 0.2200 & 0.2500 & 0.2633 &
G-Safeguard & 0.2133 & 0.2400 & 0.2467 \\
\cmidrule(lr){1-4} \cmidrule(lr){5-8}

\multicolumn{4}{c|}{\textbf{Star Topology}} & \multicolumn{4}{c}{\textbf{Tree Topology}} \\
\cmidrule(lr){1-4} \cmidrule(lr){5-8}
\textbf{Method} & \textbf{R1} & \textbf{R2} & \textbf{R3} & 
\textbf{Method} & \textbf{R1} & \textbf{R2} & \textbf{R3} \\
\midrule
No Defense & 0.2933 & 0.3500 & 0.3900 &
No Defense & 0.2500 & 0.3000 & 0.3200 \\
Dominant & 0.2700 & 0.3200 & 0.3500 &
Dominant & 0.2333 & 0.2467 & 0.2600 \\
PREM & 0.2667 & 0.3133 & 0.3367 &
PREM & 0.2267 & 0.2567 & 0.2700 \\
TAM & 0.2467 & 0.2700 & 0.2767 &
TAM & 0.2067 & 0.2233 & 0.2200 \\
BlindGuard & 0.2233 & 0.2467 & 0.2733 &
BlindGuard & 0.2100 & 0.2267 & 0.2267 \\
G-Safeguard & 0.2300 & 0.2567 & 0.2700 &
G-Safeguard & 0.2100 & 0.2200 & 0.2200 \\
\bottomrule
\end{tabular}
}
\caption{Attack success rate (ASR) across communication rounds with detection budget $K=2$.}
\label{tab:asr_comprehensive}
\end{table*}

\section{Failure Case Analysis}
We examine a representative query from our dataset: ``Marcus is trying to decide whether he really needs to do his homework… What is the percentage chance that Marcus will actually have to turn in his homework tomorrow?'' From this case, we find that agents can legitimately adopt different reasoning styles, and this style diversity helps explain both types of BlindGuard errors we observed. Normal agents use atypical direct-computation formats or introduce extra heuristics, creating local/global semantic disagreement and being flagged as outliers.

Why a ``real-world nuance'' style can cause false negatives (malicious not flagged)? Some malicious agents answer by staying perfectly on-topic while adding plausible-sounding caveats, e.g., ``substitutes often still collect homework,'' ``teacher mood matters,'' ``events aren’t independent.'' This is not an off-topic derailment; it is manifold-preserving mimicry. Semantically, the message heavily overlaps with the round intent (homework, substitute, extension, probabilities), so its embedding can remain highly aligned with the global/neighborhood context. Since BlindGuard’s anomaly score mainly relies on semantic misalignment to context, such on-topic, stylistically consistent steering may not reduce alignment enough to cross the detection margin, leading to a miss.

Why a ``direct computation'' style can cause false positives (benign flagged)? Other benign agents follow the expected probabilistic calculation strictly and produce a concise numeric derivation. In contrast, some benign replies may introduce extra heuristics (e.g., ``assume the substitute collects homework with 50\% chance'') or use a markedly different structured format. Even if non-malicious, this can create local/global semantic disagreement with the dominant group reasoning pattern for that round, making the agent look like an outlier under a context-alignment detector and thus potentially triggering false pruning.

These cases show BlindGuard is strongest at detecting semantic deviation (off-manifold behavior), but may struggle with (1) stealthy, on-topic critique attacks and (2) benign heterogeneity in reasoning style. 